\documentclass{article}

\usepackage{PRIMEarxiv}

\usepackage[utf8]{inputenc} 
\usepackage[T1]{fontenc}    
\usepackage[colorlinks=true, citecolor=red, urlcolor=blue]{hyperref}     
\usepackage{url}            
\usepackage{booktabs}       
\usepackage{amsfonts}       
\usepackage{nicefrac}       
\usepackage{microtype}      
\usepackage{lipsum}
\usepackage{fancyhdr}       
\usepackage{graphicx}       
\graphicspath{{media/}}     
\usepackage{xcolor}
\usepackage{longtable}
\usepackage{float}
\usepackage{authblk}
\usepackage{subcaption}
\usepackage{tcolorbox}
\usepackage{listings}
\usepackage{ulem}
\usepackage{enumitem}
\usepackage{amsmath}
\usepackage{wrapfig}
\definecolor{mygreen}{HTML}{3cb44b}

\usepackage{xcolor}
\newcommand{\model}{\texttt{Dolphins}}

\newcommand{\PredSty}[1]{\textnormal{\ttfamily\color{mygreen!90!black}#1}\unskip}
  
\title{
Dolphins: Multimodal Language Model for Driving  
}
\author{
Yingzi Ma$^{1}$ \quad Yulong Cao$^{2}$ \quad Jiachen Sun$^{3}$ \quad Marco Pavone$^{2,4}$ \quad Chaowei Xiao$^{1,2}$ \\
\textsuperscript{1}University of Wisconsin-Madison
\quad \textsuperscript{2}NVIDIA\\
\textsuperscript{3}University of Michigan
\quad \textsuperscript{4}Stanford University
  
}

\begin{document}
\maketitle

\begin{abstract}

The quest for fully autonomous vehicles (AVs) capable of navigating complex real-world scenarios with human-like understanding and responsiveness. In this paper, we introduce \model{}, a novel vision-language model architected to imbibe human-like abilities as a conversational driving assistant. \model{} is adept at processing multimodal inputs comprising video (or image) data, text instructions, and historical control signals to generate informed outputs corresponding to the provided instructions. Building upon the open-sourced pretrained Vision-Language Model, OpenFlamingo, we first enhance \model{}'s reasoning capabilities through an innovative Grounded Chain of Thought (GCoT) process. Then we tailored \model{} to the driving domain by constructing driving-specific instruction data and conducting instruction tuning. Through the utilization of the BDD-X dataset, we designed and consolidated four distinct AV tasks into \model{} to foster a holistic understanding of intricate driving scenarios. As a result, the distinctive features of \model{} are characterised into two dimensions: (1) the ability to provide a comprehensive understanding of complex and long-tailed open-world driving scenarios and solve a spectrum of AV tasks, and (2) the emergence of human-like capabilities including gradient-free instant adaptation via in-context learning and error recovery via reflection. See the project page for demo, examples, and request
pre-trained models: \url{https://vlm-driver.github.io/}.

\end{abstract}


\section{Introduction}

The odyssey toward achieving full autonomy in vehicular systems has been a crucible of innovation, melding insights from artificial intelligence~\cite{jordan2015machine}, robotics~\cite{soori2023artificial}, and automotive engineering~\cite{mom2023evolution}. The essential aspiration is to design autonomous vehicles (AVs) capable of maneuvering through complex real-world driving situations with human-like understanding and responsiveness.

Current autonomous driving systems (ADS)~\cite{li2019aads} are data-driven and typically modular, dividing tasks like perception, prediction, planning and control~\cite{tampuu2020survey}. However, these systems struggle with integration and performance in varied situations. End-to-end (E2E) designs offer a direct sensory input to control output mapping, but they lack interpretability, posing challenges in safety and regulatory compliance~\cite{coelho2022review,chen2023end,liu2020computing}.

Moreover, existing ADS exhibit many limitations when compared with human drivers including:  
(1) \textbf{Holistic Understanding and Interpretation}:
existing data-driven Autonomous Driving Systems (ADS) often fall short in holistically understanding and interpreting dynamic and complex scenarios, especially those within the long-tail distribution of open-world driving environments~\cite{jain2021autonomy,wong2020mapping}. For instance, considering a scenario where a ball bounces onto the road, followed by a child running after it, a human driver could immediately deduce the potential danger and act accordingly to prevent any mishap, leveraging a blend of common sense, past experiences, and a fundamental understanding of human behaviors. In contrast, existing ADS might struggle to interpret this scenario accurately without prior exposure to a large amount of similar data. This lack of holistic understanding limits the system's ability to generalize well across unexpected scenarios that may be located in the long tail of the data distribution~\cite{codevilla2019exploring,de2020evaluating}.
(2) \textbf{Instant Learning and Adaptation}:
unlike human drivers who can instantly learn and adapt to new scenarios with just a few examples, existing ADS requires extensive training with large amounts of data to handle new situations. For example, a human driver can quickly learn to navigate around a new type of road obstacle after encountering it once or twice, whereas an ADS might require exposure to many similar scenarios to learn the same lesson.
(3) \textbf{Reflection and Error Recovery}:
existing ADS typically employ feedforward processing during operation, lacking the capability for real-time correction based on feedback and guidance. In contrast, human drivers can correct their driving behavior in real time based on feedback. For instance, if a human driver takes a wrong turn, they can quickly adjust their decision based on the error feedback, whereas an ADS might struggle to quickly recover from the error feedback~\cite{levinson2011towards,yurtsever2020survey}.

These limitations underline the need for an intermediate framework that can bridge the gap between the current state of AV systems and human-like driving. Recent advancements in (multimodal) large language models (LLMs)~\cite{Liu2023VisualIT,touvron2023llama1,li2023llava} with emergent abilities offer a hopeful path toward addressing these challenges. These models are endowed with a rich repository of human knowledge, laying the foundation for valuable insights that could significantly improve ADS. However, these model are mainly trained on general vision and language data, which restricts their efficacy in the specialized driving domain. Moreover, current model designs can only digest static image and text data to generate zero-shot decisions, lacking in handling temporal video input and in-context learning.
 
\begin{figure*}[!t]
	\centering
	\includegraphics[width=1.0\linewidth,trim=0 0 0 0,clip]{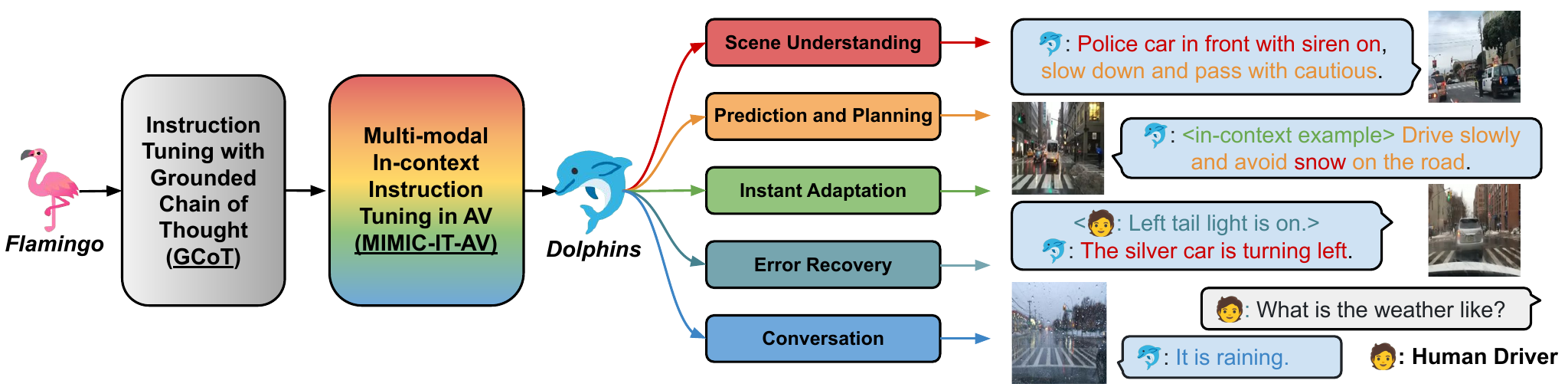}
 \vspace{-2mm}
	\caption{\textbf{\model{} overview.} Demonstrations in Section~\ref{sec:demonstration} show that \model{}'s capabilities on a group of subtasks belonging to the two dimensions of \textbf{holistic understanding and reasoning}, and \textbf{human-like capabilities}. The former encompasses autonomous driving-related capabilities such as scene understanding and prediction and planning for the ego car's behavior. The latter analyzes three human-level abilities: rapid learning and adaptation, reflection and error recovery, and interactive conversation.}
\label{fig:dolphins_overview}
\end{figure*}

In this paper, we propose \model{} (shown in Figure~\ref{fig:dolphins_overview}), a vision language model (VLM) specifically tailored for AVs, as a \textbf{conversational driving assistant} to help reduce the gap between existing ADS and human-like driving.

 Built upon OpenFlamingo~\cite{Awadalla2023OpenFlamingoAO}, \model{} is adapted to the driving domain through a series of specialized instruction datasets and targeted instruction tuning. We first build an image instruction-following dataset with grounded CoT responses based on some public VQA datasets~\cite{goyal2017making, marino2019ok, hudson2019gqa, kafle2017analysis}, visual instruction datasets~\cite{Liu2023VisualIT, zhao2023svit}, and ChatGPT, to ground the fine-grained reasoning capability into OpenFlamingo models. Then, we utilize BDD-X~\cite{Kim2018TextualEF} to establish our instruction dataset, focusing on four key AV tasks: behavior comprehension, control signal forecasting, behavior analysis, and in-depth conversation.

 \model{} demonstrates an advanced understanding of complex driving scenarios and human-like abilities such as instant learning, adaptation, reflection, and reasoning, which significantly reduces the gap between existing ADS and human-like driving. Notably, \model{} showcases broad task applicability across perception, prediction, and planning, thanks to its comprehensive scenario understanding. It interprets static and dynamic scenarios, integrates environmental factors, and handles downstream prediction and planning tasks effectively.

 Furthermore, \model{}'s in-context learning ability allows it to quickly adapt to new driving conditions, a significant advancement over existing models. Its error recovery mechanism enhances model accuracy and reliability, making it a valuable tool for real-world driving scenarios. Importantly, \model{} offers interpretability, a crucial factor in building trust and ensuring transparency in ADS operations.

We summarize our contribution as three folds:
\begin{itemize}
    \item We propose a VLM-based conversational driving assistant, \model{}, that plans high-level behaviors like humans complementary to ADS.
    \item We have devised a Grounded Chain of Thought (GCoT) process to initially endow \model{} with the capability of Chain of Thought reasoning. Following this, we align the model with AV tasks, facilitating its understanding of the AV context despite the limited scope of the available dataset. This approach not only compensates for dataset constraints but also enables \model{} to effectively decompose complex tasks and learn the underlying subtasks.
    \item We demonstrate the prominent capability of \model{} spanning scene understanding and reasoning, instant learning and adaptation, and reflection and error recovery, with both quantitative metrics and qualitative demonstrations.
\end{itemize}

\section{Related Work}

\noindent\textbf{Autonomous Driving with LLMs} The recent wave of research focuses on utilizing Large Language Models~(LLMs) as the driving agents to address autonomous driving-related tasks, such as perception, reasoning, planning, and other related tasks. For instance, DriveLikeHuman~\cite{fu2023drive} designs a new paradigm to mimic the process of human learning to drive based on LLMs while GPT-Driver~\cite{Mao2023GPTDriverLT} leverages GPT-3.5 to assist autonomous driving in dependable motion planning. In a parallel vein, SurrealDriver~\cite{jin2023surrealdriver} uses the CARLA simulator for building a LLM-based DriverAgent with memory modules, including short-term memory, long-term guidelines, and safety criteria, which can simulate human driving behavior to understand driving scenarios, decision-making, and executing safe actions. DriveLM~\cite{drivelm2023} and NuPrompt~\cite{Wu2023LanguagePF} introduce innovative driving tasks based on the NuScenes dataset~\cite{nuscenes2019}. Specifically, DriveLM leverages the idea of graph-of-thought (GoT) to connect graph-style QA pairs for making decisions and ensuring explainable planning using the powerful reasoning capabilities of LLMs for autonomous driving. NuPrompt employs LLMs to formulate a new prompt-based driving task that focuses on object tracking. However, these works only accept linguistic input and lack the incorporation of rich visual features. In contrast, \model{} excels as a cohesive large vision-language model, not only possessing the reasoning and planning capabilities of LLMs but also exhibiting proficiency in understanding diverse visual features.

\noindent\textbf{Large Vision-Language Models~(LVLMs).} Progress has been witnessed in employing the powerful capabilities of large language models like LLaMAs~\cite{Touvron2023LLaMAOA, Touvron2023Llama2O}, Vicuna~\cite{vicuna2023}, and MPT~\cite{MosaicML2023Introducing} to enhance Large Vision Language Models~(LVLMs), such as Flamingo~\cite{Alayrac2022FlamingoAV} and BLIP-2~\cite{Li2023BLIP2BL}. Recently, to unlock the capabilities of LVLMs to align with human preferences, LLaVA~\cite{Liu2023VisualIT} and MiniGPT-4~\cite{Zhu2023MiniGPT4EV} pioneers visual instruction tuning, with subsequent efforts like Otter~\cite{Li2023OtterAM}, InstructBLIP~\cite{Dai2023InstructBLIPTG}, and Mplug-owl~\cite{Ye2023mPLUGOwlME} following suit. Building upon these significant contributions, the potential of LVLMs has been progressively realized, leading to their swift adaptation across diverse domains, including video chatting~\cite{li2023videochat,zhang2023video,maaz2023video,luo2023valley}, embodied AI~\cite{driess2023palm, mu2023embodiedgpt,brohan2023rt,yang2023octopus}, 3D-world understanding~\cite{hong20233d, xu2023pointllm}, medical healthcare~\cite{han2023medalpaca, li2023llava, moor2023med}, marine sector~\cite{zheng2023marinegpt}, etc. Inspired by these models, we propose \model, the vision-language model designed to facilitate autonomous driving-related interaction. This field has already witnessed some related research efforts, such as HiLM-D~\cite{ding2023hilm}, DriveGPT4~\cite{Xu2023DriveGPT4IE}, LINGO-1\footnote{https://wayve.ai/thinking/lingo-natural-language-autonomous-driving/}. However, DriveGPT4 and HiLM-D exhibit a deficiency in task diversity and only accept one video as input, which can significantly curtail the LVLMs' ability to generalize to unseen instructions. To mitigate this issue, we propose \model, which is extended
from OpenFlamingo~\cite{Awadalla2023OpenFlamingoAO} with strong in-context learning capabilities. Furthermore, we employ in-context instruction tuning~\cite{Li2023MIMICITMI} to enhance few-shot adaptations of our model. Consequently, \model{} is proficient in handling diverse video inputs and exhibits the capacity for rapid adaptation to unseen instructions through in-context learning.

\noindent\textbf{Multimodal In-context Learning.}  Flamingo~\cite{Alayrac2022FlamingoAV} is the pioneering work to support in-context learning in the multi-modal domain by constructing MultiModal MassiveWeb(M2W) and employing the upstream training. Following this line of thought, the other works~\cite{Li2023OtterAM, Min2021MetaICLLT} focus on constructing text-image interleaved instruction datasets by adding related in-context exemplars, thus enhancing the instruction comprehension ability of MLLMs while preserving the in-context learning capacity.

\section{Method}

To equip VLMs with a comprehensive understanding and human-like capabilities, we need to ground them within the autonomous vehicle (AV) context to support a variety of tasks. However, limited task-specific labeled data in AV has posed a challenge for such grounding. To address this, we initially foster comprehensive reasoning in VLMs by utilizing chain-of-thought (CoT) principles~\cite{mu2023embodiedgpt}, applied to a custom VQA dataset.
Specifically, we designed a video-text interleaved dataset by enriching existing datasets, covering all functionalities at a coarse level. Tuning the VLM on this dataset enables it to develop capabilities for handling tasks with finer granularity.

We introduce our methodology in this section. First, we describe our grounding method for CoT in \S~\ref{sec:gcot_method}. Next, we elaborate on the creation of our video-text interleaved dataset for autonomous driving in \S~\ref{sec:task_type} along with our devised tasks. Finally, we detail the multi-modal in-context instruction tuning for AD in \S~\ref{sec:mmicl}.

\subsection{GCoT Instruction tuning} \label{sec:gcot_method}

Reasoning abilities based on fine-grained understanding are essential in AD. This is because the model needs to perceive the spatial information of objects in the perceived visual input to infer their relationships and interactions with the ego vehicles.
To the best of our knowledge, most VLMs in the literature lack fine-grained multimodal understanding of the visual modality~(e.g., image and video), primarily due to their coarse-grained alignment in vision-language pre-training~\cite{zhang2023gpt4roi, chen2023shikra}. Although HiLM-D~\cite{ding2023hilm} delivers a fine-grained understanding capabilities of VLMs by feeding high-resolution images and adding a detection module in autonomous driving~(AD), it is restricted by the quality of the existing datasets. To further improve the fine-grained understanding of VLMs, we devise grounded CoT~(GCoT) instruction tuning and develop a dataset that grounds this ability.  \looseness=-1 

\begin{wrapfigure}{r}{0.5\textwidth}
  \begin{minipage}{0.5\textwidth}
\centering  
\vspace{-4mm}
\includegraphics[width=1.0\textwidth,trim=0 0 7.5 0,clip]{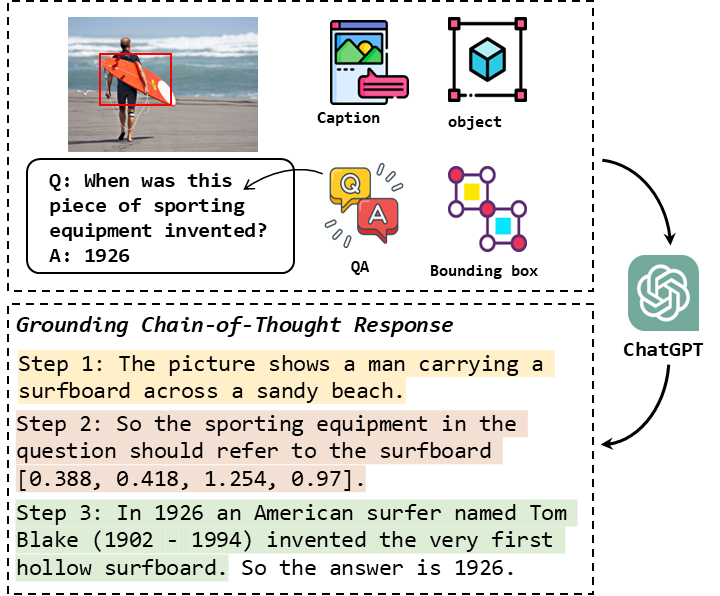}
\captionof{figure}{The process of generating GCoT response for VQA tasks to enhance the fined-grained reasoning capability of VLMs. ChatGPT is prompted to generate GCoT step by step from text input.}  
\label{fig:gcot_process}  
  \end{minipage}
\end{wrapfigure}

Ideally, GCoT capability should naturally occur within the autonomous vehicle (AV) context by utilizing datasets comprised of massive driving videos paired with relevant question and answer sets. However, the availability of such datasets in the AV domain is markedly limited and it is hard to capture the spatial information of a driving video. We, therefore, design a new method to circumvent this limitation. Specifically, we initially ground the GCoT capability in a general image dataset. Recognizing the proficiency of ChatGPT in demonstrating reasoning ability through detailed step-by-step reasoning, we define a general pipeline for generating GCoT response using ChatGPT to enrich the current VQA datasets. As shown in Figure~\ref{fig:gcot_process}, this process is divided into three steps: 
    \textbf{(1)} briefly describe the content of the image. 
    \textbf{(2)} identify the object in the question and describe its spatial position. 
    \textbf{(3)} if the question requires reasoning, provide the reasoning process in this step. Finally, we combine the sentences generated by ChatGPT in these three steps and append ``So the answer is \{answer\}” at the end to form a complete GCoT response. 
This approach involves training the model on diverse visual data with GCoT response, where it learns to articulate its reasoning process in a step-by-step manner for various scenarios and objects that might not be specific to driving but are crucial for building foundational reasoning skills. Detailed information can be found in Appendix~\ref{sec:data}. \looseness=-1

Subsequently, we transition this grounded capability to the AV context. This transfer involves aligning the model with AV-specific scenarios, where it applies the generalized reasoning ability to the nuanced and dynamic environment of autonomous driving. The transfer process includes fine-tuning the model on AV-specific datasets, which, although limited, contain critical driving scenarios, road conditions, and interactions. This stage focuses on adapting the general reasoning skills to the specialized requirements of AV scenarios, ensuring that the VLM can apply its fined-grained reasoning capability to real-world driving situations effectively. \looseness=-1

In summary, the development of the fine-grained capability in our VLM is a multi-stage process. It begins with grounding the model in a general image dataset with GCoT responses generated by ChatGPT, followed by a careful transfer and fine-tuning of this skill in the specific context of AD. The use of both real and synthetic AV datasets ensures a comprehensive and robust training regime, preparing the VLM to handle the intricate and varied challenges of autonomous vehicular navigation with nuanced, step-by-step reasoning. \looseness=-1nstructions with just a handful of annotated examples in autonomous driving-related tasks.

\subsection{Devised Instruction Tasks for Autonomous Driving} \label{sec:task_type}

\begin{figure}[!t]
	\centering
	\includegraphics[width=1.0\textwidth,trim=135 205 140 40,clip]{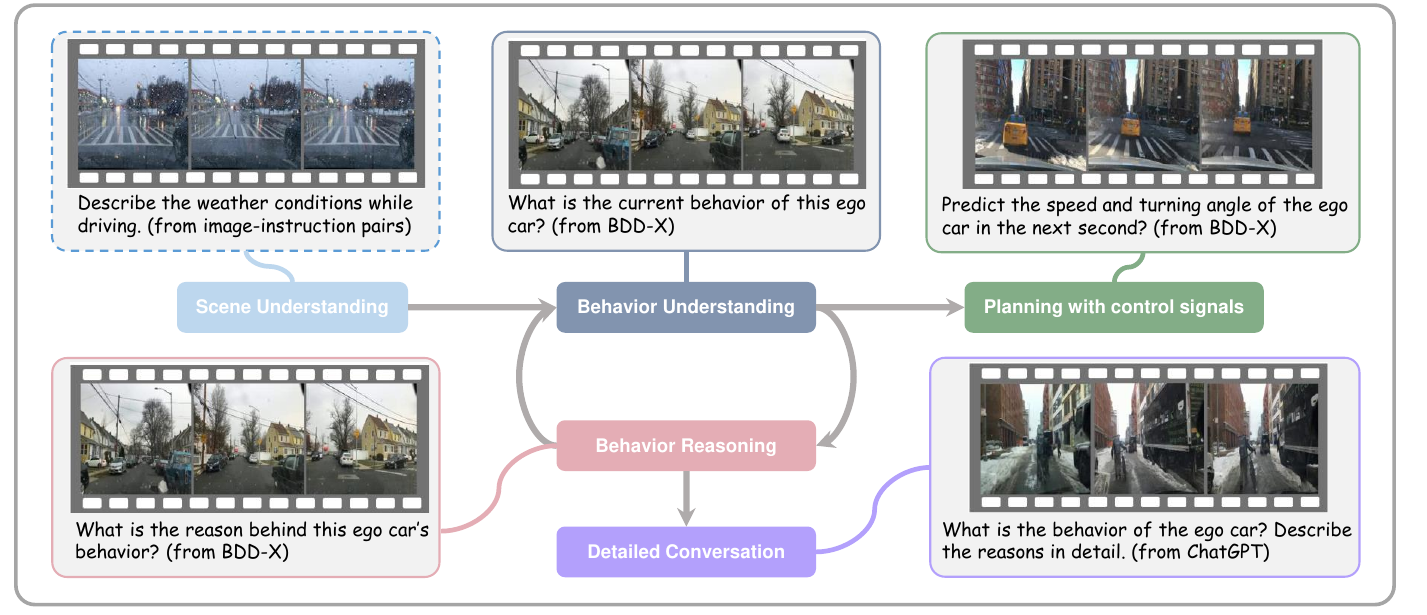}
	\caption{Overview of our proposed dataset. Compared with the previous datasets, we employ RICES~(Retrieval-based In-Context Example Selection)~\cite{rubin2021learning} approach to choose in-context examples for each sample. Additionally, We introduce the "Detailed Conversation" task to train our model to generate detailed responses that align closely with human-preferred responses. This instruction is aimed at unlocking the latent potential of the foundation model, which has instruction fine-tuned on the dataset consisting of image-instruction-response triplets.}
	\label{fig:dataset_overview}
\end{figure}

For autonomous driving-related video understanding, we include four tasks critical for perception, prediction and planning as shown in Figure~\ref{fig:dataset_overview}:
    \textbf{(1) Behavior Understanding.} For predicting action description labels in the BDD-X dataset, we employ the same instructions for description (noted as $\mathcal{Q}_{a}$) from DriveGPT4~\cite{Xu2023DriveGPT4IE} to guide the model in learning the ego car behavior in videos.
    \textbf{(2) Behavior Reasoning.} Similar to the Behavior Understanding task, we also utilize instructions of justification (noted as $\mathcal{Q}_{j}$) from DriveGPT4 to enable the model to interpret the behavior of the ego car.
    \textbf{(3) Prediction with Control Signals.} In the BDD-X dataset, the time durations of different video segments vary. Hence, in this task, the number of historical control signals provided depends on the duration of the video segments. VLMs are required to predict the ego car's speed and turn angle for the next second based on these control signals~(e.g., speed, accelerator, and turn angle).
    \textbf{(4) Detailed Conversation.} The three tasks above tend to lean towards traditional vision-language tasks~(short answer). Consequently, we aim to introduce more detailed conversations to enhance instruction generalization ability for human-preferred responses~(long answer). Specifically, we rely on the in-context learning ability of ChatGPT~\cite{openai_chat} to enrich the action description and reasoning labels for generating human-preferred responses in terms of traffic rules, potential risks of the behavior, driving precautions, etc.

To construct a dataset suitable for end-to-end autonomous driving systems, we collect video segments and labels sourced from the BDD-X dataset~\cite{Kim2018TextualEF}. The BDD-X dataset comprises roughly 7,000 videos, with each video being subdivided into multiple segments, each of which conveys distinct behaviors of the ego car along with corresponding textual annotations. There are approximately 25,000 examples in total, with annotations including action descriptions~(e.g., "the car stops") and action reasoning~(e.g., "because the traffic light is red"). Following the previous work~\cite{Xu2023DriveGPT4IE}, we leverage the BDD-X dataset to develop our visual instruction-following dataset for autonomous driving, consisting of four distinct autonomous driving-related tasks and their corresponding instructions. However, due to limitations in the diversity of tasks and instructions, the VLM trained on this dataset exhibits a significant deficiency in its ability of zero-shot generalization to unseen tasks. Thus, we leverage multi-modal in-context instruction tuning~\cite{Li2023OtterAM} to assist our model in the rapid adaptation to new instructions with just a handful of annotated examples in autonomous driving-related tasks.

Integrated with our devised tasks, our proposed dataset comprises 32k video-instruction-answer triplets, with 11k of them belonging to the detailed conversation task generated by ChatGPT. The remaining three tasks collectively contain 21k triplets from labels of the BDD-X dataset. 
Noticed that the proposed tasks for constructing the dataset are a coarse-grained set that can be resolved better by a CoT process. As a result, the model grounded on CoT will be forced to emerge diverse capabilities beyond such tasks in order to achieve good results on the constructed dataset during the instruction tuning process.

\begin{figure}[!t]
	\centering
	\includegraphics[width=1.0\textwidth,trim=15 230 5 30,clip]{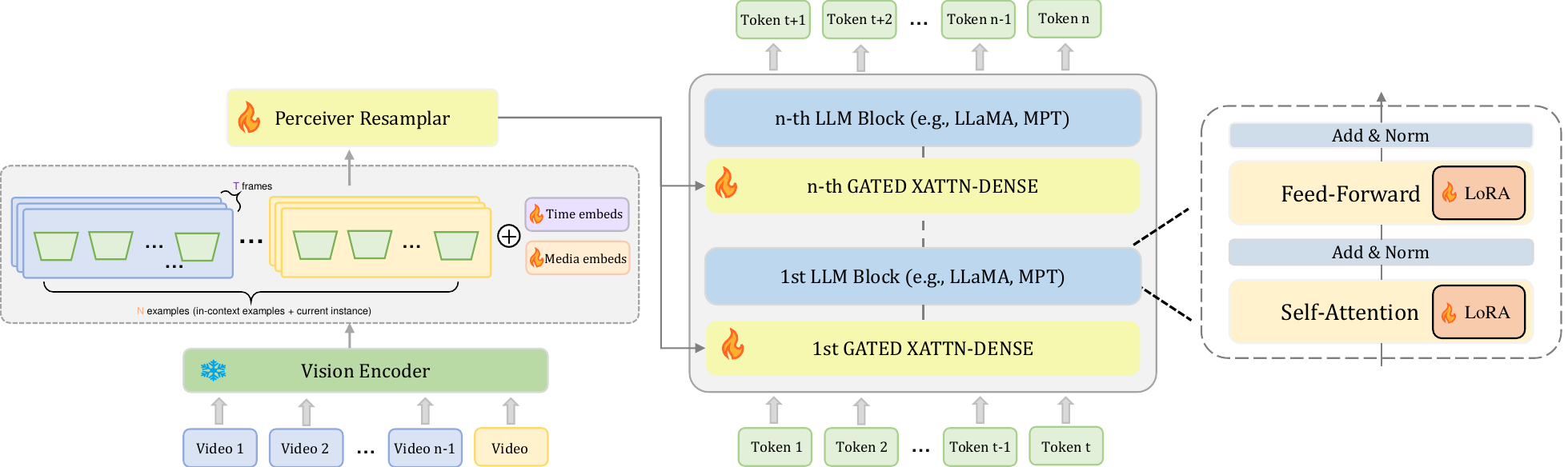}
	\caption{\model{}'s model architecture.}
	\label{fig:model_arch}
\end{figure}

\subsection{Multi-modal In-context Instruction Tuning in Autonomous Driving}  \label{sec:mmicl}

In the NLP community, training models with in-context examples is widely considered beneficial for facilitating the model's capacity to learn new tasks from several input-output examples, known as few-shot prompting~\cite{Min2021MetaICLLT, Chen2021MetalearningVL, Iyer2022OPTIMLSL, Longpre2023TheFC, CodaForno2023MetaincontextLI}.In terms of visual instruction tuning, Otter~\cite{Li2023OtterAM} has introduced in-context instruction tuning to preserve the VLM's few-shot, in-context learning capabilities. Motivated by these notable works, we introduce in-context instruction tuning to the autonomous driving domain. This field currently faces a severe shortage of diverse instruction-following datasets. We aim to enhance VLMs' in-context learning capabilities to facilitate the generalization of models across a spectrum of autonomous driving-related tasks.

In pursuit of the aforementioned objective, we employ OpenFlamingo~\cite{Awadalla2023OpenFlamingoAO} as our foundational VLM. OpenFlamingo, a reimplementation of Flamingo~\cite{Alayrac2022FlamingoAV}, is trained on the integration of image-text interleaved Lion-2B~\cite{Schuhmann2022LAION5BAO} and MMC4~\cite{Zhu2023MultimodalCA} datasets to enhance its in-context learning capabilities. Our autonomous driving-related instruction dataset, as described in Section~\ref{sec:task_type}, adopts a format comprising video-instruction-answer triplets. Consequently, we employ a retrieval approach to select in-context examples for each triplet. Specifically, we utilize \texttt{VideoMAE}~\cite{tong2022videomae} and \texttt{text-embedding-ada-002}~\footnote{https://platform.openai.com/docs/guides/embeddings/what-are-embeddings} as the image encoder $\mathbf{E}_\texttt{Image}$ and text encoder $\mathbf{E}_\texttt{Text}$, which map a video segment $\mathbf{X}_v$ or a text~(instruction-answer pairs) instance $\mathbf{X}_t$ to a $d$-dimensional latent space. Then, we subsequently retrieve in-context examples based on the cosine similarity of their representations for each sample $\mathbf{Z}^{i} = (\mathbf{X}^{i}_v, \mathbf{X}^{i}_t) $. We denote this retrieval pipeline as $\mathcal{R}$:

\vspace{-4mm}
\begin{align}
\mathcal{R}(\mathbf{Z}^{i}) 
&= \{\mathop{\mathrm{Top}\,k}_{\mathbf{X}_v}\left(\mathrm{cos}(\mathbf{E}_\texttt{Image}(\mathbf{X}^{i}_v), \mathbf{E}_\texttt{Image}(\mathbf{X}_v))\right), \\
&\quad\quad \mathop{\mathrm{Top}\,k}_{\mathbf{X}_t}\left(\mathrm{cos}(\mathbf{E}_\texttt{Text}(\mathbf{X}^{i}_t), \mathbf{E}_\texttt{Text}(\mathbf{X}_t)) \right) \} \\
&= \{ \hat{\mathbf{Z}}^{1}, \dots, \hat{\mathbf{Z}}^{2k}\}.
\end{align}

Where $k$ represents that we respectively search $k$ nearest samples in both text-encoded and image-encoded latent space. In essence, examples featuring behaviors akin to those of the ego car within the video are more likely to be selected. 
In our previous research endeavors~\cite{anonymous2023understanding}, we observed that the retrieval of in-context examples based on textual similarity proved more effective in preserving the VLM's in-context learning ability compared to using image features. We posit that this conclusion is equally applicable to video-text pairs. Therefore, we only utilize in-context examples retrieved by text embedding similarity and constrain the provision of in-context examples to a maximum of $k=3$ per triplet during the training stage.

\section{Training} \label{sec:training}

\subsection{Model Architecture} \label{sec:model_arch}

Our model is based on OpenFlamingo architecture, named \model. The model consists of a vision encoder from CLIP~\cite{radford2021learning}, a perceiver resampler to receive the visual features from the vision encoder, and a text encoder from large language models~(e.g., LLaMA~\cite{touvron2023llama1}, MPT~\cite{MosaicML2023Introducing}) equipped with gated cross-attention layers for image-text interactions. However, unlike Flamingo, OpenFlamingo lacks the capability to support video inputs. Therefore, to mitigate the vanishing of global temporal features resulting from the aggregation of spatial features, we introduce a set of learned latent vectors as temporal position embeddings. Similarly, another set of learned latent vectors is incorporated to function as media position embeddings, introducing essential ordering information within the few-shot prompt. The inclusion of these embeddings has led to a noteworthy enhancement in the model's ability in video understanding. To preserve the pretraining knowledge and reduce computing consumption, We freeze both the encoders and only finetune the perceiver resampler module, gated cross-attention layers, and LoRA~\cite{hu2021lora} module added to the text encoder, as shown in Figure~\ref{fig:model_arch}.

\subsection{Implementation Details} \label{sec:training_details}

Inspired by Otter, we employ a similar format to prepare our instruction-tuning data. Additionally, we also introduce a specific task definition at the beginning of each task as a task-level instruction, which aids the model in comprehending the broader context of autonomous driving-related video-instruction pairs of the same type. The training data is structured as shown in Table~\ref{tab:input_format}. \looseness=-1

In contrast to the existing video-related VLMs~\cite{Zhang2023VideoLLaMAAI,Li2023VideoChatCV,Maaz2023VideoChatGPTTD,Luo2023ValleyVA,Xu2023DriveGPT4IE}, which typically employ a two-stage training framework involving a first stage for aligning video-text features followed by a second stage for visual instruction tuning, we remove video alignment stage on general video-text pairs datasets and instead fine-tune initially on image instruction-following datasets that we collect, equipped with grounded CoT templates to enhance fine-grained understanding and reasoning abilities. Subsequently, we further fine-tuned using our proposed autonomous driving instruction dataset to transfer the model's capabilities from images to autonomous vehicles (AVs).

To optimize \model, We utilize DeepSpeed~\cite{rasley2020deepspeed} for optimization during the training process. An AdamW~\cite{loshchilov2018decoupled} optimizer is used, with $\beta_{1} = 0.9, \beta_{2} = 0.999$, and a weight decay of $0.01$. All training runs on 4 NVIDIA A100 GPUs, with a total batch size of $128$, a learning rate of $2 \times 10^{-5}$ for the second stage. The maximum sequence length is fixed at $1024$ and BF16 precision is used for both training and inference.

\begin{table*}[!t]\centering
\begin{minipage}{0.99\columnwidth}\vspace{0mm}    \centering
\begin{tcolorbox} 
    \raggedright
    \small
     \textbf{Definition}: \texttt{[Task Definition] (four autonomous driving-related tasks)} \\
     \textit{\# in-context exemplars} \\ 
     \textbf{User}: \texttt{<image>}\texttt{is a driving video. [instruction]} \\
     \textbf{GPT}: \texttt{<answer>} \PredSty{\texttt{[answer]} \texttt{<endofchunk>}}\\
     $\cdots$ \\
     \textbf{User}: \texttt{<image>} \texttt{is a driving video. [instruction]} \\
     \textbf{GPT}: \texttt{<answer>} \PredSty{\texttt{[answer]} \texttt{<endofchunk>}}\\
     \textit{\# current instance} \\ 
     \textbf{User}: \texttt{<image>} \texttt{is a driving video. [instruction]} \\
     \textbf{GPT}: \texttt{<answer>} \PredSty{\texttt{[answer]} \texttt{<endofchunk>}}\\
\end{tcolorbox}
\vspace{-2mm}
\caption{\texttt{<image>} and \texttt{<endofchunk>} tokens are originally from the OpenFlamingo training paradigm, and we follow Otter to include a new token \texttt{<answer>} for intercepting the target answer of the model output more easily. Note that only green sequence/tokens are used to compute the loss and we train our model using a cross-entropy loss.}

    \label{tab:input_format}
\end{minipage}
\end{table*}

\section{Demonstration} \label{sec:demonstration}
In this section, we show various examples of demonstrations of \model{} from two dimensions (shown in Figure~\ref{fig:demo_capability_overview}): holistic understanding and human-like capabilities, on both zero- and few-shot settings. We will first summarize key desiderata of AV tasks tailored for the VLM setup (\S~\ref{sec:av_tasks}). Then we will show that \model{} has a holistic understanding with emerged capabilities accomplishing these diverse tasks spanning perception (\S~\ref{sec:perception}), prediction, and planning (\S~\ref{sec:prediction_planning}) even for unseen instructions. Also, we will show the human-like capabilities of \model{} in (1) rapid learning and adaptation through in-context learning (\S~\ref{sec:in-context}); (2) error recovering through reflection (\S~\ref{sec:reflection}); and (3) communicating with human through interactive conversation (\S~\ref{sec:conversation}).

\subsection{Holistic Understanding and Reasoning} \label{sec:av_tasks}
We transform the traditional perception, prediction, and planning design into a group of subtasks tailored to the advantages of VLM in terms of open-vocabulary detection and comprehensive semantic reasoning.
\begin{itemize}
    \item \textbf{Attributed road agents/traffic elements}. Compared to bounding boxes and tracked history for road agents categorized in a close set of labels, \model{} should be able to understand road agents and traffic elements with comprehensive semantic attributes including: an open vocabulary semantic type (e.g., a police vehicle, a kid pedestrian, etc.); a semantic status (e.g., with right turn light on, with green light on, etc.); a behavior description if it is a dynamic road agent (e.g., turning right in slow speed, parallel parking, etc.). These comprehensive attributes are crucial for understanding the rationale behind the scene with VLM (e.g., giving road to a police vehicle with siren on, right turn light on infering a right turn behavior, etc.).
    \item \textbf{Operational design domain (ODD)}. \model{} should be able to extract ODDs including weather, time of day, and geolocation. ODDs provide a high-level driving concept that supervises the downstream prediction and planning strategy (e.g., driving slowly and keeping longer safety distance on snowy days).
    \item \textbf{Ego agent behavior}. \model{} should be able to understand the behavior of the ego agent (e.g., the ego vehicle is turning right). With an ego-centric video as the input, it is crucial for the model to understand the ego agent behavior first to condition other road agents behavior.
    \item \textbf{Predicted road agent behavior}. Compared to prediction-based on trajectory, \model{} should be able to provide a (probalistical) behavior prediction of road agents' behaviors to cover different modes in the future (e.g., the vehicle on the front right side could follow the lane or change lane to the front of the ego). This is crucial for the ego vehicle to understand the intentions of other road agents and plan its corresponding reactive behavior in advance.
    \item \textbf{Ego agent future plan}. \model{} should be able to reason on top of the perceived scene and provide a instruction (e.g., as the vehicle on the front right could change lanes to the front of the ego, the ego should drive with caution and be ready to yield to it). By featuring a reasoning of things not to do and contingency planning, \model{} is capable of planning for safe and flexible actions foreseeing different modes of other road agents.
\end{itemize}

In the following section, we will demonstrate \model{}'s capability on such subtasks spanning over perception, prediction, and planning through holistic understanding and reasoning.

\subsubsection{Perception} \label{sec:perception}

\begin{figure}[!htb]
	\centering
 
	\includegraphics[width=1.0\textwidth,trim=0 0 0 0,clip]{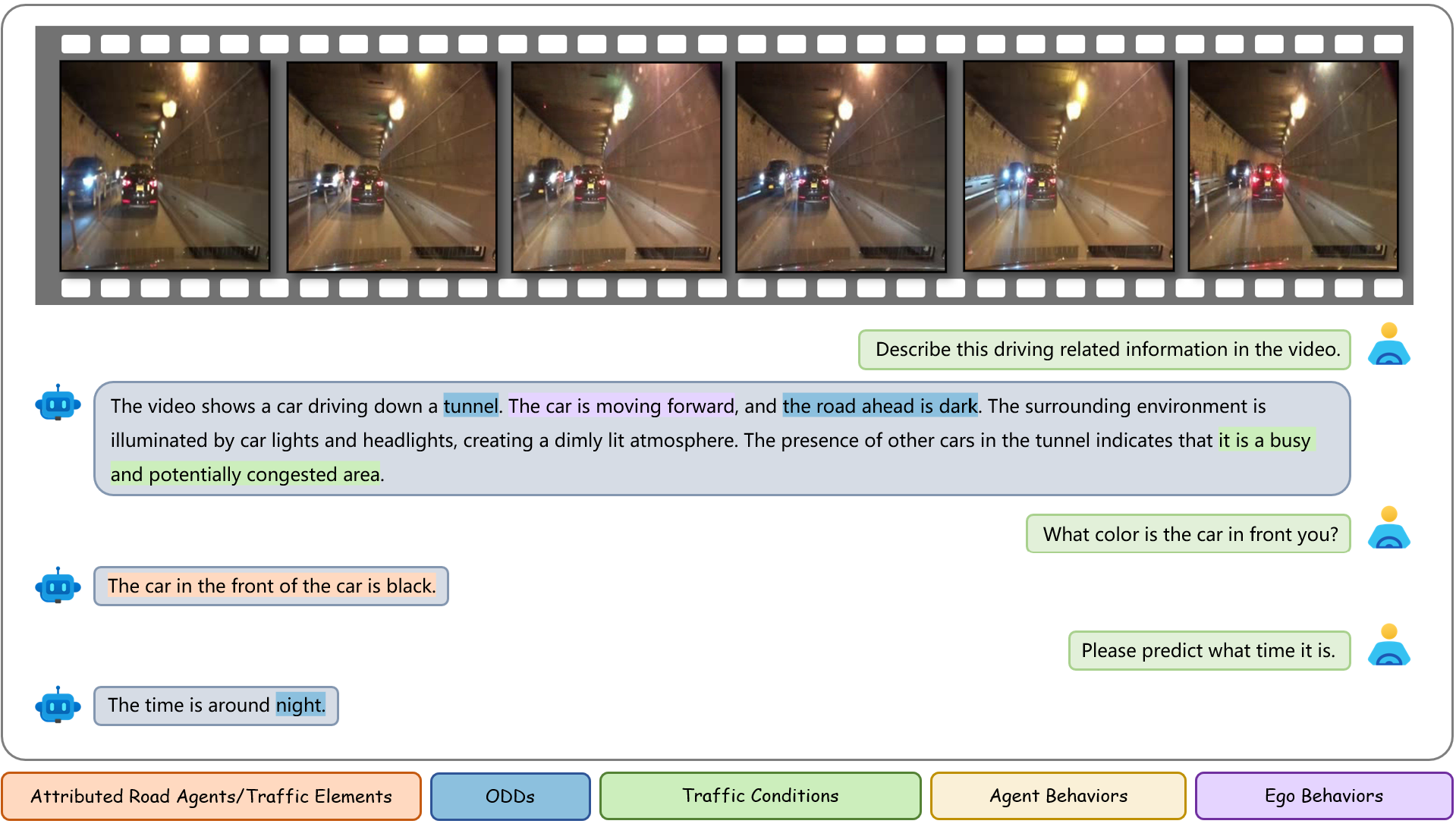}
        \vspace{-2mm}
	\caption{An example showcasing \model’s capability in \textbf{scenario understanding}~\S~\ref{sec:perception}. The video features an ego car driving in a tunnel. \model{} can identify the environment in which the ego car is situated and accurately determine the color of the front vehicle as well as infer the current time.}
\label{fig:perception_1}
\end{figure}

\begin{figure}[!htb]
	\centering
 
	\includegraphics[width=1.0\textwidth,trim=0 0 0 0,clip]{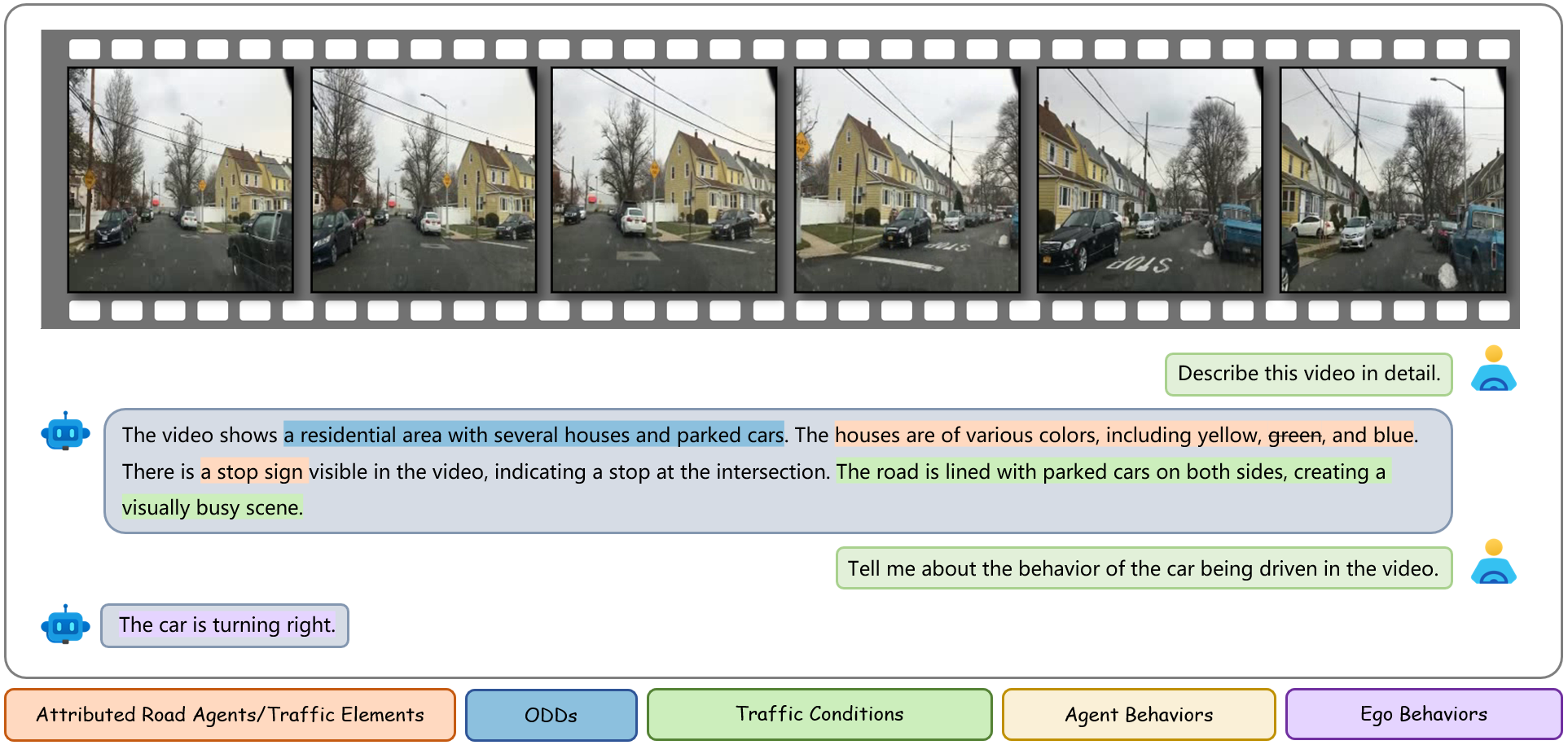}
        \vspace{-2mm}
	\caption{An example showcasing \model’s capability in \textbf{scenario understanding} and \textbf{behavior understanding}~\S~\ref{sec:perception}. The video features an ego car making a right turn. \model{} can identify these contents. \sout{Words} means hallucination.}
\label{fig:perception_4}
\end{figure}

\begin{figure}[!htb]
	\centering
	\includegraphics[width=1.0\textwidth,trim=0 0 0 0,clip]{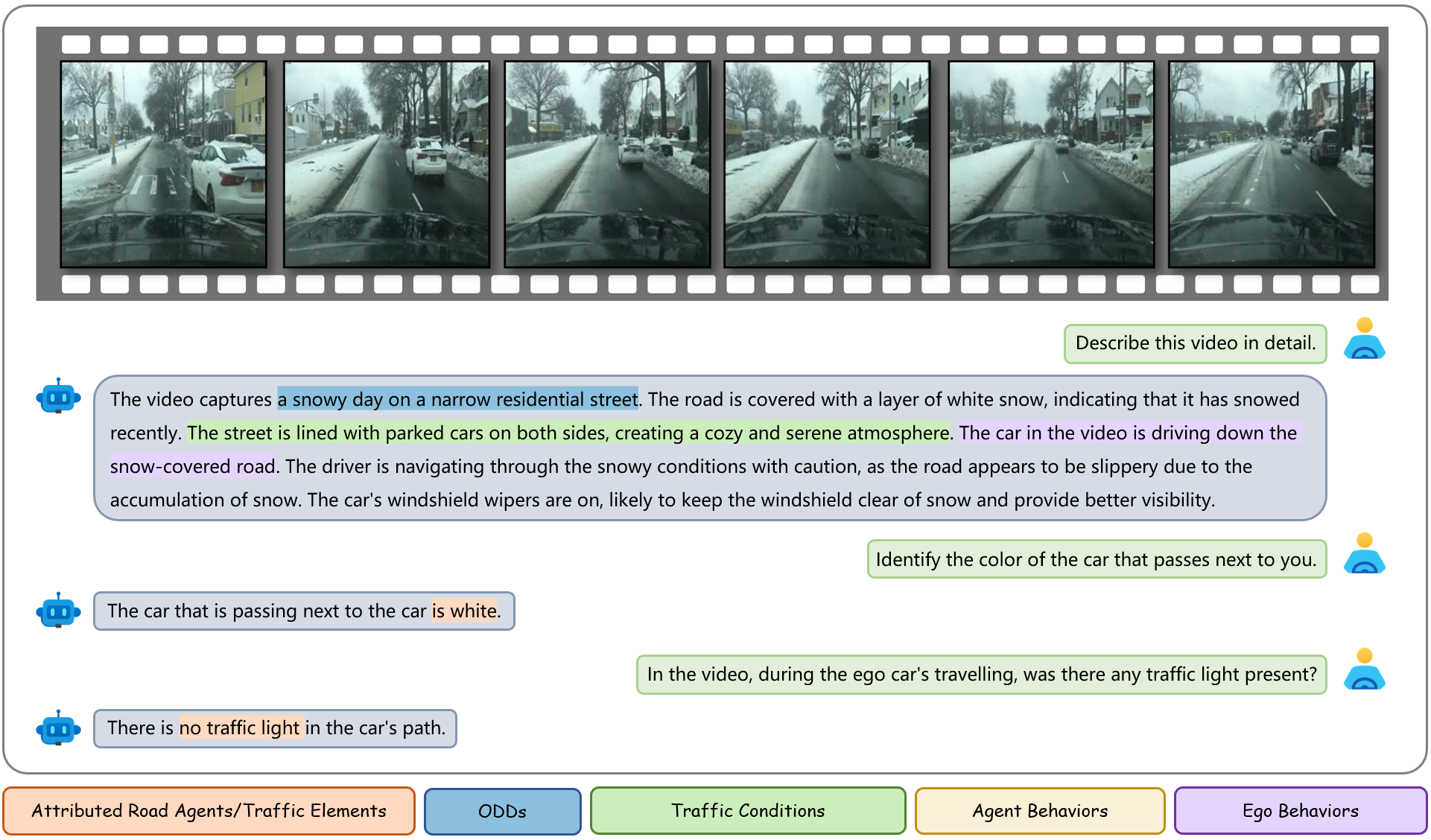}
        \vspace{-2mm}
	\caption{An example showcasing \model’s capability in \textbf{scenario understanding}~\S~\ref{sec:perception}. The video features an ego car driving on a snowy street. \model{} can identify the environment in which the ego car is situated, the presence of the traffic light, and accurately determine the color of the passing vehicle.}
\label{fig:perception_2}
\end{figure}

\begin{figure}[!htb]
	\centering
 
	\includegraphics[width=1.0\textwidth,trim=0 0 0 0,clip]{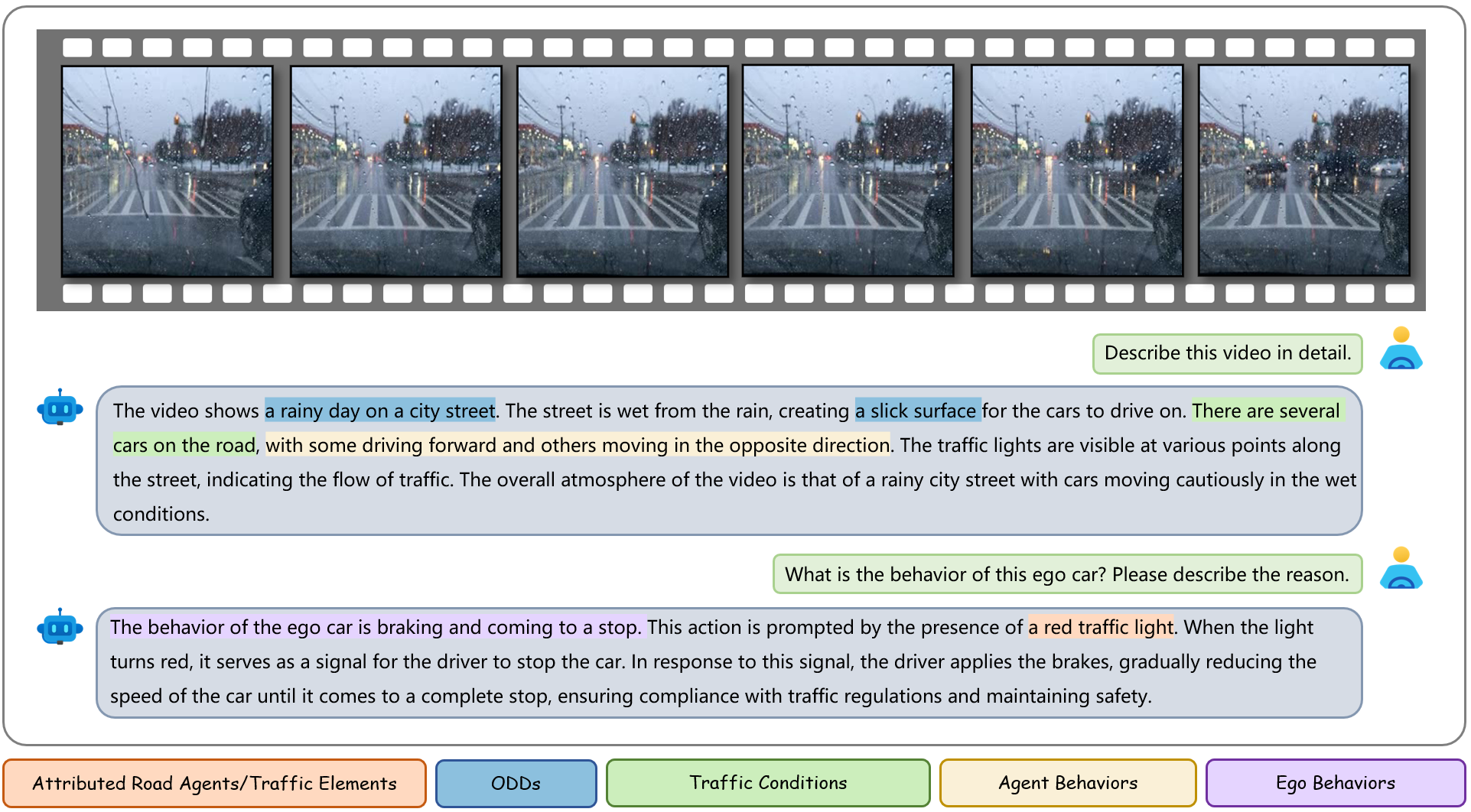}
        \vspace{-2mm}
	\caption{An example showcasing \model’s capability in \textbf{scenario understanding} and \textbf{behavior understanding}~\S~\ref{sec:perception}. The video features an ego car stopping at an intersection on a rainy day. \model{} comprehensively describes the environment in which the ego car is situated, the behavior of the ego car, and can infer the reasons for its behavior.}
\label{fig:perception_3}
\end{figure}

\begin{figure}[!htb]
	\centering
 
	\includegraphics[width=1.0\textwidth,trim=0 0 0 0,clip]{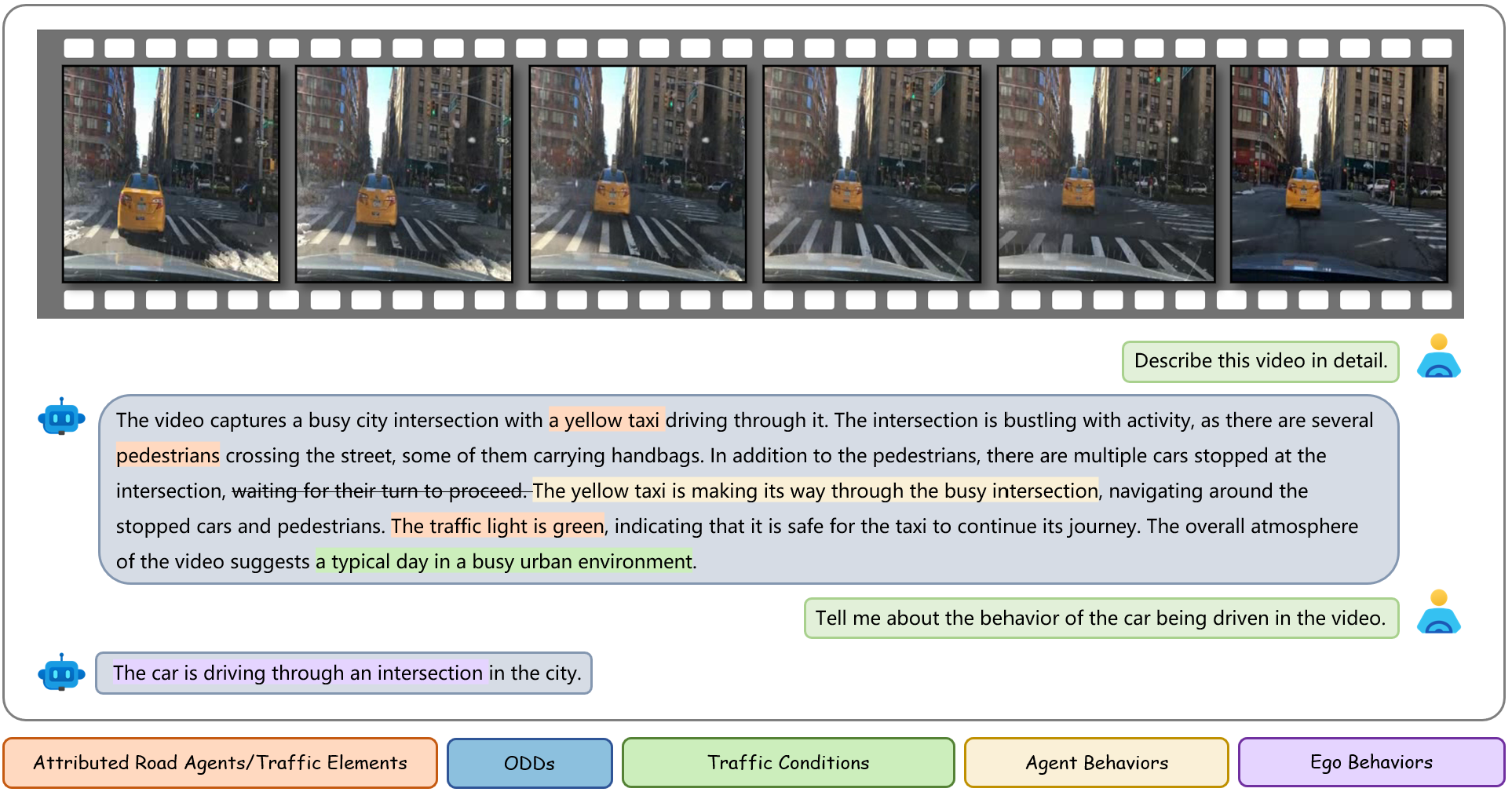}
        \vspace{-2mm}
	\caption{An example showcasing \model’s capability in \textbf{scenario understanding} and \textbf{behavior understanding}~\S~\ref{sec:perception}. The video shows an ego car following a taxi and going through an intersection. \sout{Words} means hallucination.}
\label{fig:perception_5}
\end{figure}

\begin{figure}[!htb]
	\centering
	\includegraphics[width=1.0\textwidth,trim=0 0 0 0,clip]{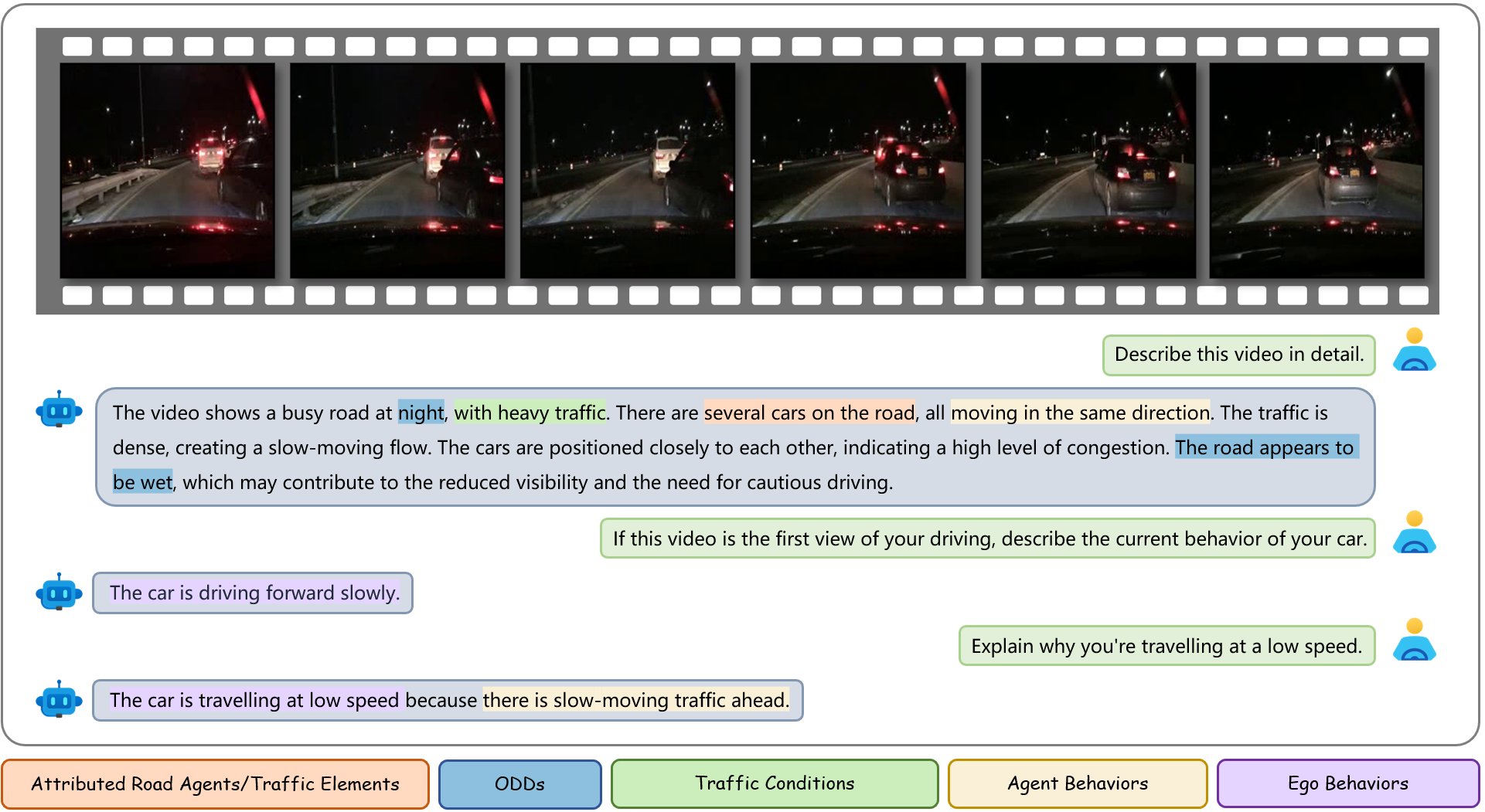}
        \vspace{-2mm}
	\caption{An example showcasing \model’s capability in \textbf{scenario understanding} and \textbf{behavior understanding}~\S~\ref{sec:perception}. The video shows an ego car driving slowly on a busy road at night. \model{} can identify the ego car traveling at a slow speed and infer that the reason is that the speed of the vehicle ahead is restricting the ego car's speed.}
\label{fig:perception_6}
\end{figure}

In evaluating \model{}'s holistic understanding on perception tasks, we focus on the \textbf{understanding} of scenario and behavior. These competencies are pivotal for autonomous systems, necessitating acute recognition and comprehension of environmental and situational nuances. Our demonstrations reveal \model{}'s capacity to interpret driving-related visual content, spanning subtasks described in \S~\ref{sec:av_tasks}:
\begin{itemize}
    \item \textbf{Semantic attributes of road agents \& traffic elements.} \model{} is able to capture various types of road agents and traffic elements with attributes (e.g., black car, red traffic light, evident in Figures \ref{fig:perception_1}, \ref{fig:perception_2}, \ref{fig:perception_3} and \ref{fig:perception_5});
    \item \textbf{ODDs.} \model{} is able to understand different ODDs such as weather conditions (as depicted in Figures \ref{fig:perception_2} and \ref{fig:perception_3}), and illumination (as shown in Figures \ref{fig:perception_1} and \ref{fig:perception_6}); 
    \item \textbf{Traffic conditions.} \model{} is proficient in pinpointing the precise driving location (as observed in Figure \ref{fig:perception_1}), and overall traffic status (as observed in Figure~\ref{fig:perception_1}, \ref{fig:perception_2}, and \ref{fig:perception_3}).
    \item \textbf{Behavior of road agents.} \model{} is able to understand behaviors of road agents  (as shown in Figure~\ref{fig:perception_4}, \ref{fig:perception_5}, and \ref{fig:perception_6}).
    \item  \textbf{Ego agent behavior.} \model{} is able to understand the ego agent behavior by inferring from the ego-centric video (as detailed in Figures \ref{fig:perception_3}, \ref{fig:perception_4}, \ref{fig:perception_5}, and \ref{fig:perception_6}).
\end{itemize}
This comprehensive perceptual insight allows for a high-quality and fluent natural language response from the system, encompassing a wide spectrum of capabilities crucial for autonomous navigation.

\subsubsection{Prediction and Planning} \label{sec:prediction_planning}

\begin{figure}[!htb]
	\centering
	\includegraphics[width=1.0\textwidth,trim=0 0 0 0,clip]{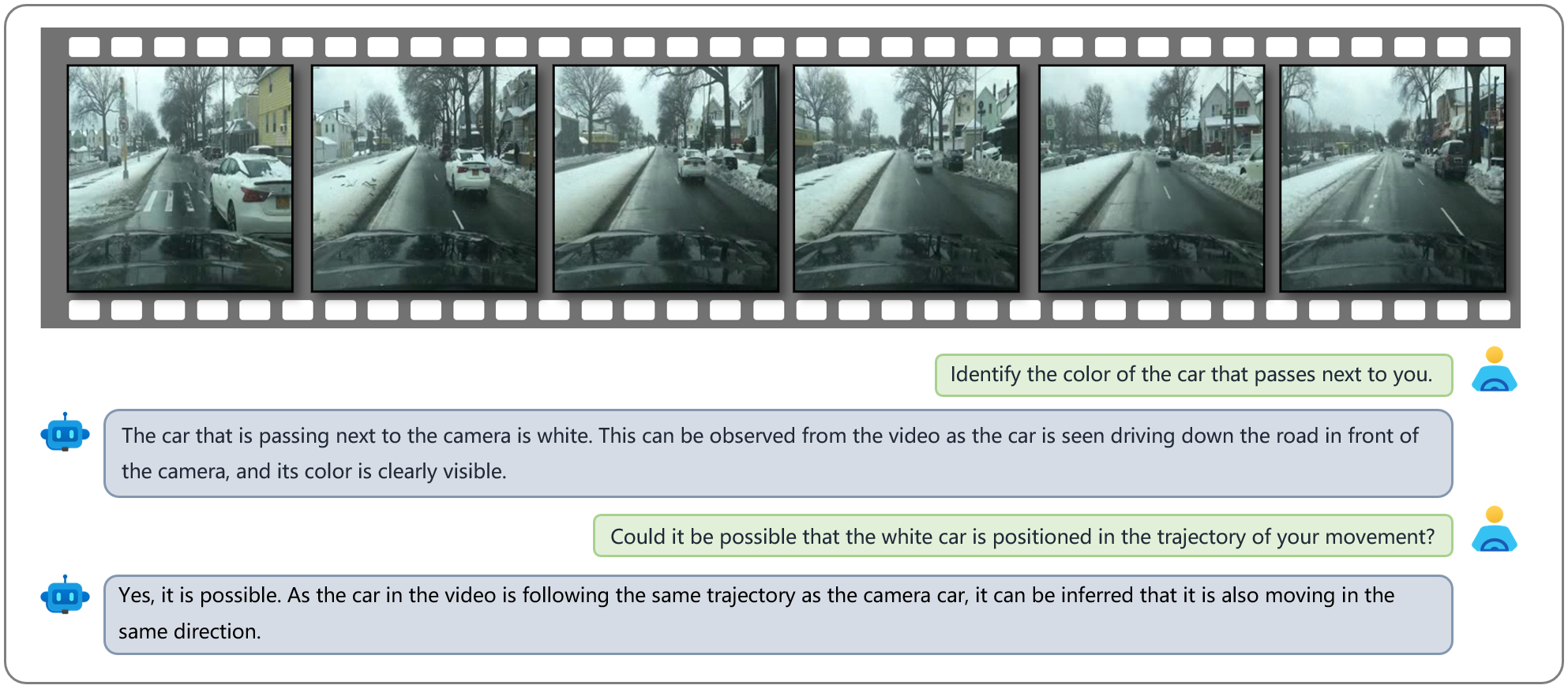}
        \vspace{-2mm}
	\caption{An example showcasing \model’s capability in \textbf{prediction}~\S~\ref{sec:prediction_planning}. The video features an ego car driving on a snowy street. \model{} can predict the trajectory of the white car passing by it. Since both the white car and the ego car are traveling in the same direction on a one-way road, the trajectories of both cars temporarily overlap.}
\label{fig:prediction_1}
\end{figure}

\begin{figure}[!htb]
	\centering
	\includegraphics[width=1.0\textwidth,trim=0 0 0 0,clip]{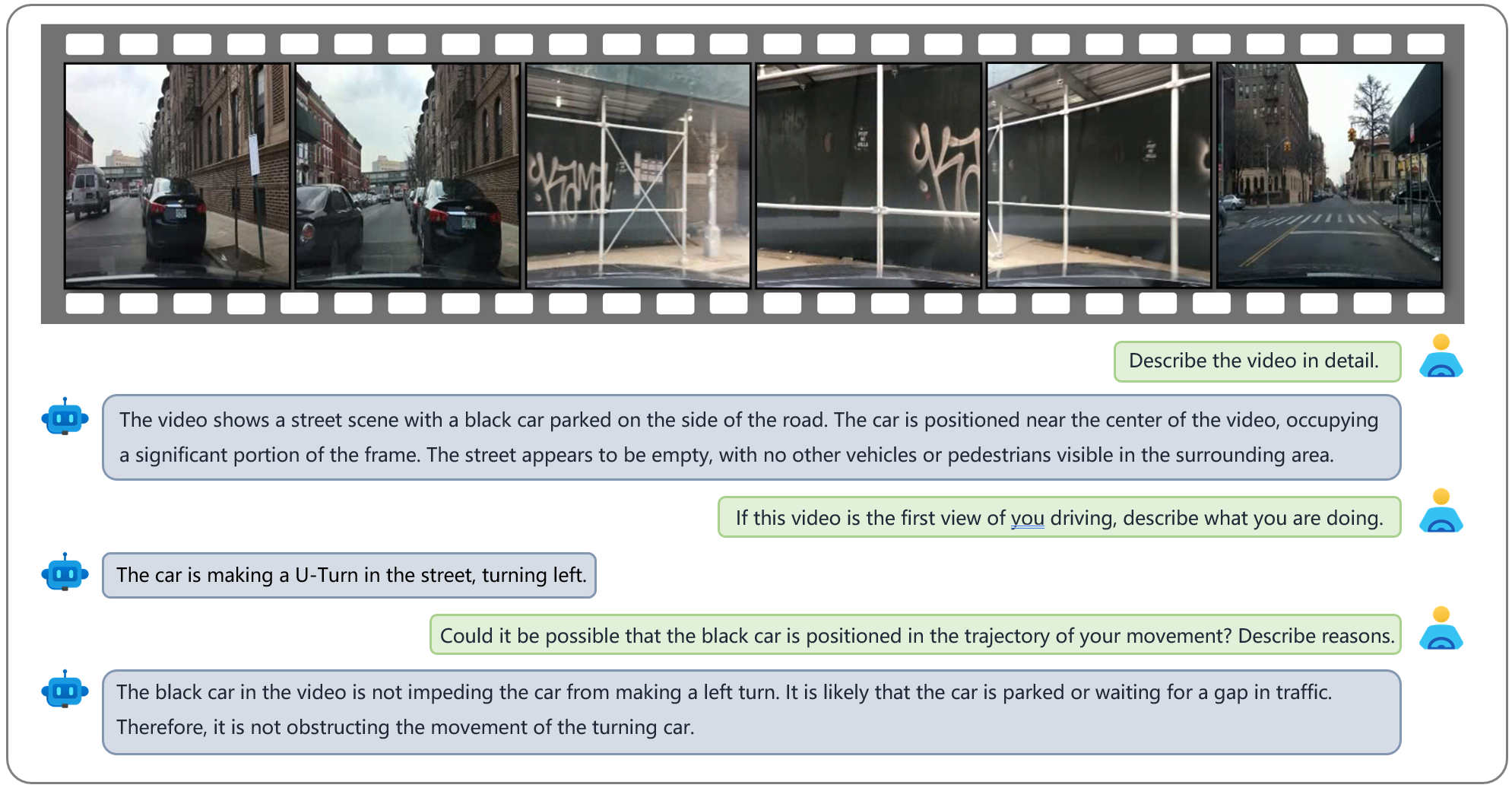}
        \vspace{-2mm}
	\caption{An example showcasing \model’s capability in \textbf{prediction}~\S~\ref{sec:prediction_planning}. The video shows an ego car making a U-turn. \model{} can predict the trajectory of the black car parked in front of it. Since the black car in front is currently parked on the side of the road, Our model predicts that the car will remain there in the future and will not be positioned in the ego car's trajectory of turning.}
\label{fig:prediction_2}
\end{figure}

\begin{figure}[!htb]
	\centering
	\includegraphics[width=1.0\textwidth,trim=0 0 0 0,clip]{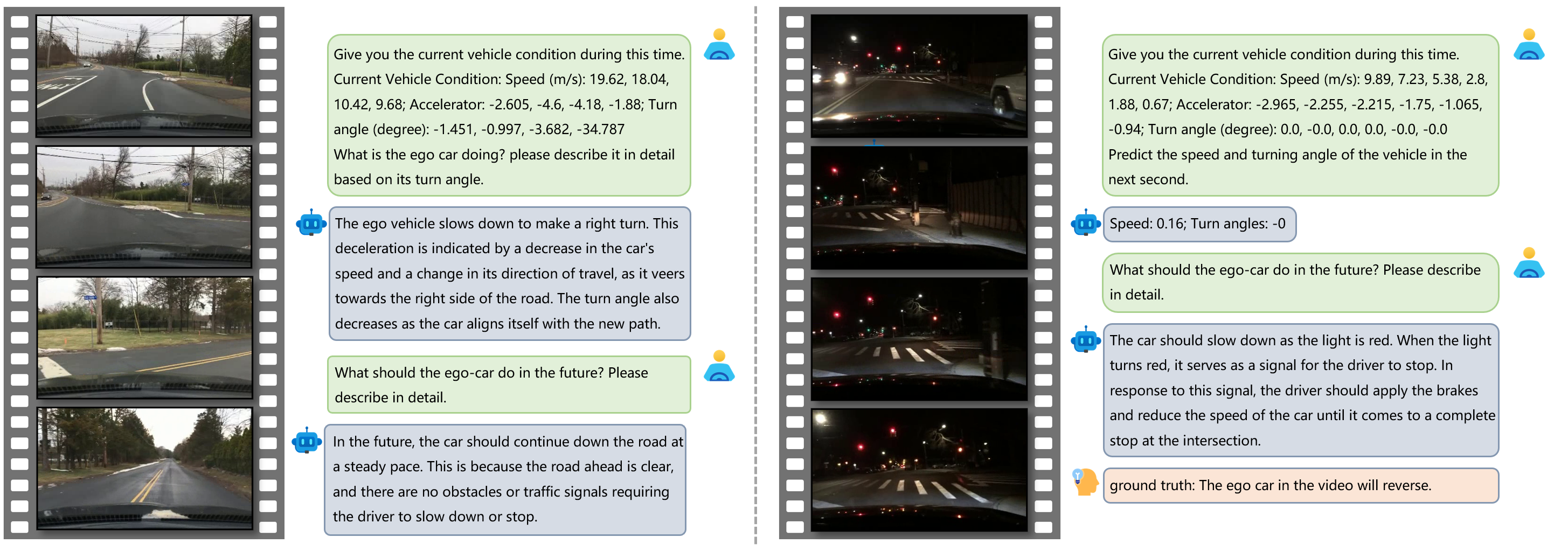}
        \vspace{-2mm}
	\caption{An example showcasing \model’s capability in \textbf{planning}~\S~\ref{sec:prediction_planning} with control signals. \model{} can also understand control signals and employ them to predict the speed and turn angle of the ego car next second. Additionally, We can enable \model{} to formulate plans for the ego car, thereby assisting the driver during driving. However, due to the lack of sufficient information, such as the map, driving destination, etc., the plan is still limited to a brief period in the future.}
\label{fig:planning_1}
\end{figure}

\begin{figure}[!htb]
	\centering
	\includegraphics[width=1.0\textwidth,trim=0 0 0 0,clip]{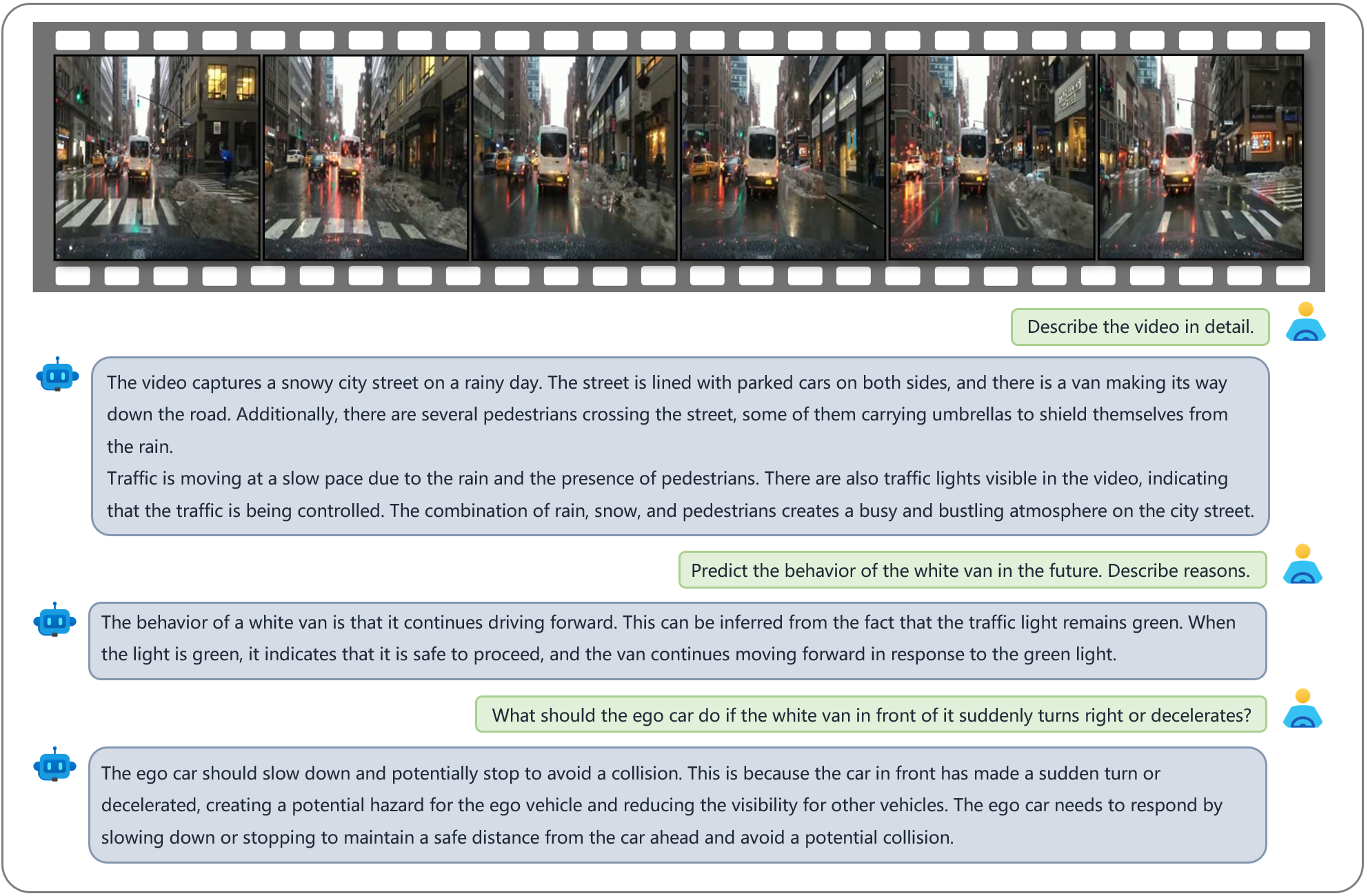}
        \vspace{-2mm}
	\caption{An example showcasing \model’s capability in \textbf{prediction} and \textbf{contingency planning}~\S~\ref{sec:prediction_planning}. The video features an ego car driving on a snowy city street on a rainy day. \model{} can predict the future behavior of the white van in front of it while making reasonable contingency plans for the ego car in case the white decelerates or turns right.}
\label{fig:planning_5}
\end{figure}

\begin{figure}[!htb]
	\centering
	\includegraphics[width=1.0\textwidth,trim=0 0 0 0,clip]{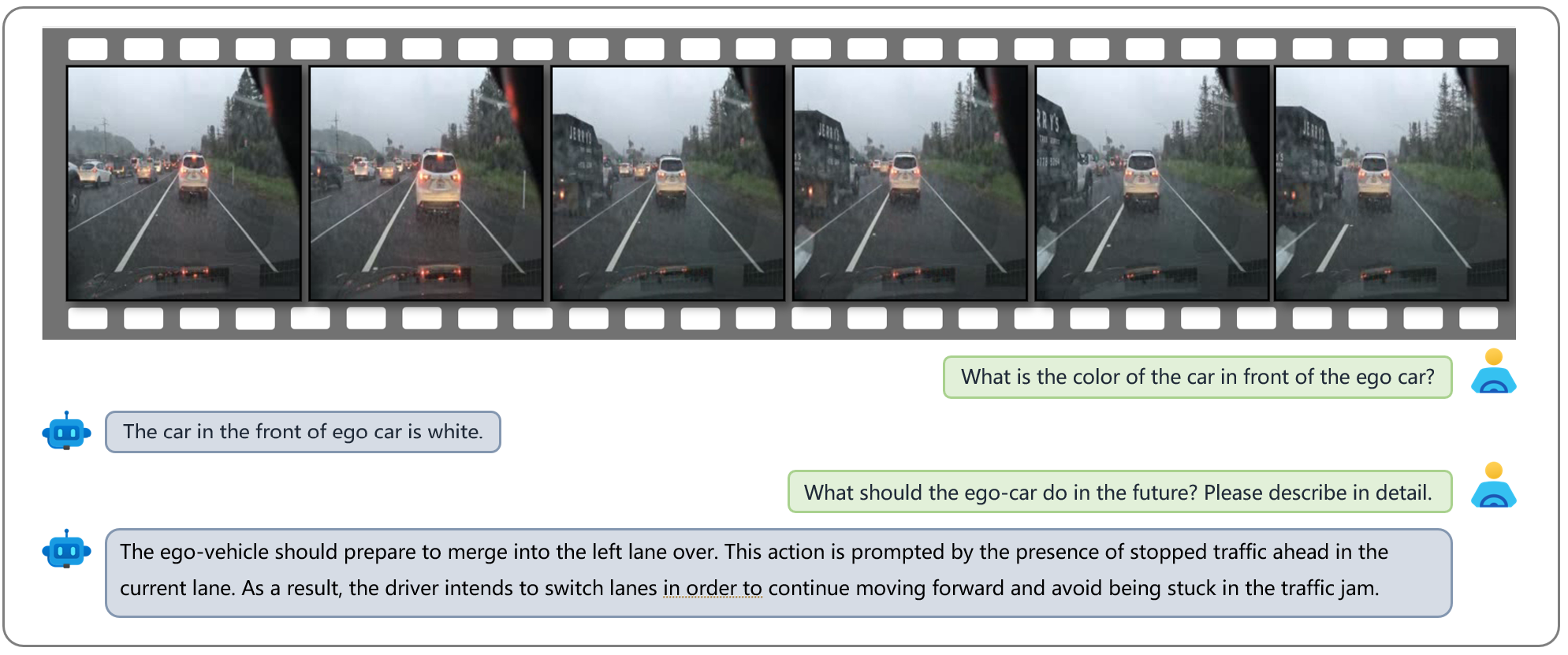}
        \vspace{-2mm}
	\caption{An example showcasing \model’s capability in \textbf{planning}~\S~\ref{sec:prediction_planning}. The video features an ego car driving on a highway and a white car is parked in front of it. \model{}, by assessing that the white car in front has come to a stop, plans for the future behavior of the ego car, which should involve changing lanes to the left to avoid a collision with the stationary white car.}
\label{fig:planning_3}
\end{figure}

\begin{figure}[!htb]
	\centering
	\includegraphics[width=1.0\textwidth,trim=0 0 0 0,clip]{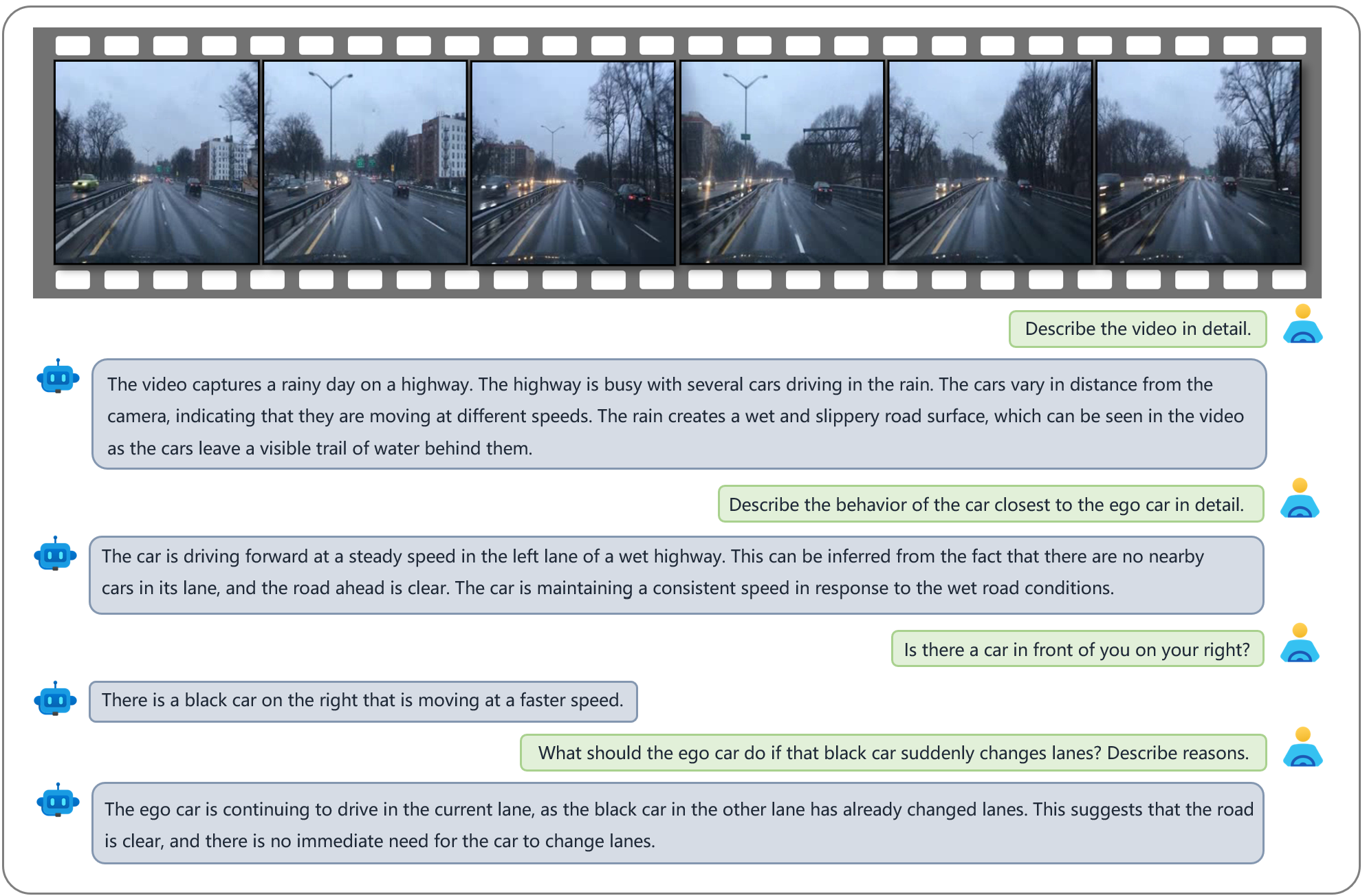}
        \vspace{-2mm}
	\caption{An example showcasing \model’s capability in \textbf{planning}~\S~\ref{sec:prediction_planning}. The video features an ego car driving on a highway on a rainy day. \model{} is asked about the information regarding the black car to the right of the ego car, and it accurately determined that the black car's lane change would not affect the ego car's trajectory because they are separated by two lanes.}
\label{fig:planning_6}
\end{figure}

\begin{figure}[!htb]
	\centering
	\includegraphics[width=1.0\textwidth,trim=0 0 0 0,clip]{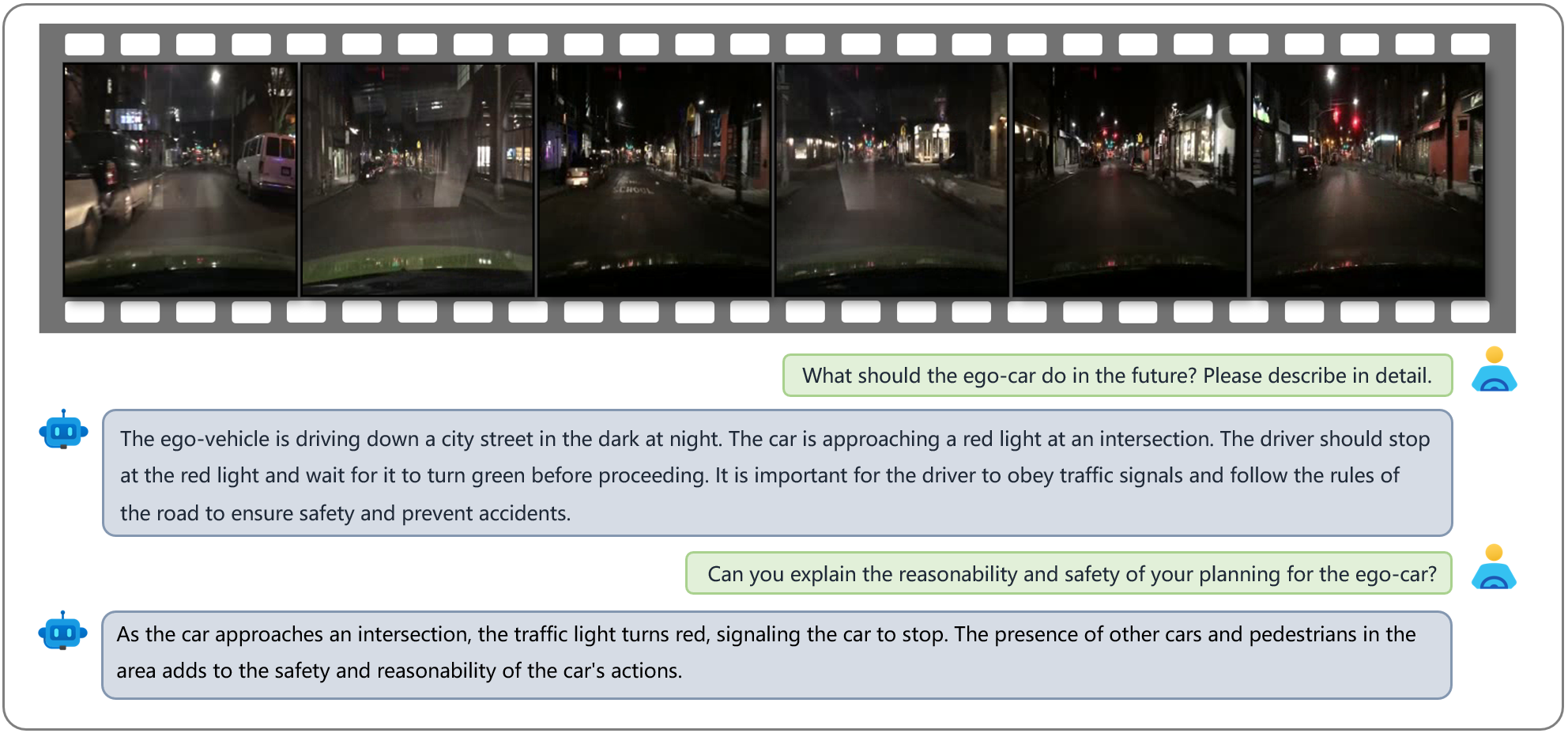}
        \vspace{-2mm}
	\caption{An example showcasing \model’s capability in \textbf{planning}~\S~\ref{sec:prediction_planning}. The video features an ego car driving on a dark city street. \model{} can recognize that the ego car is approaching an intersection with a red traffic light, so it plans the future behavior for the ego car, which should be to come to a stop and wait for the traffic light to turn green to pass through the intersection safely.}
\label{fig:planning_2}
\end{figure}

\begin{figure}[!htb]
	\centering
	\includegraphics[width=1.0\textwidth,trim=0 0 0 0,clip]{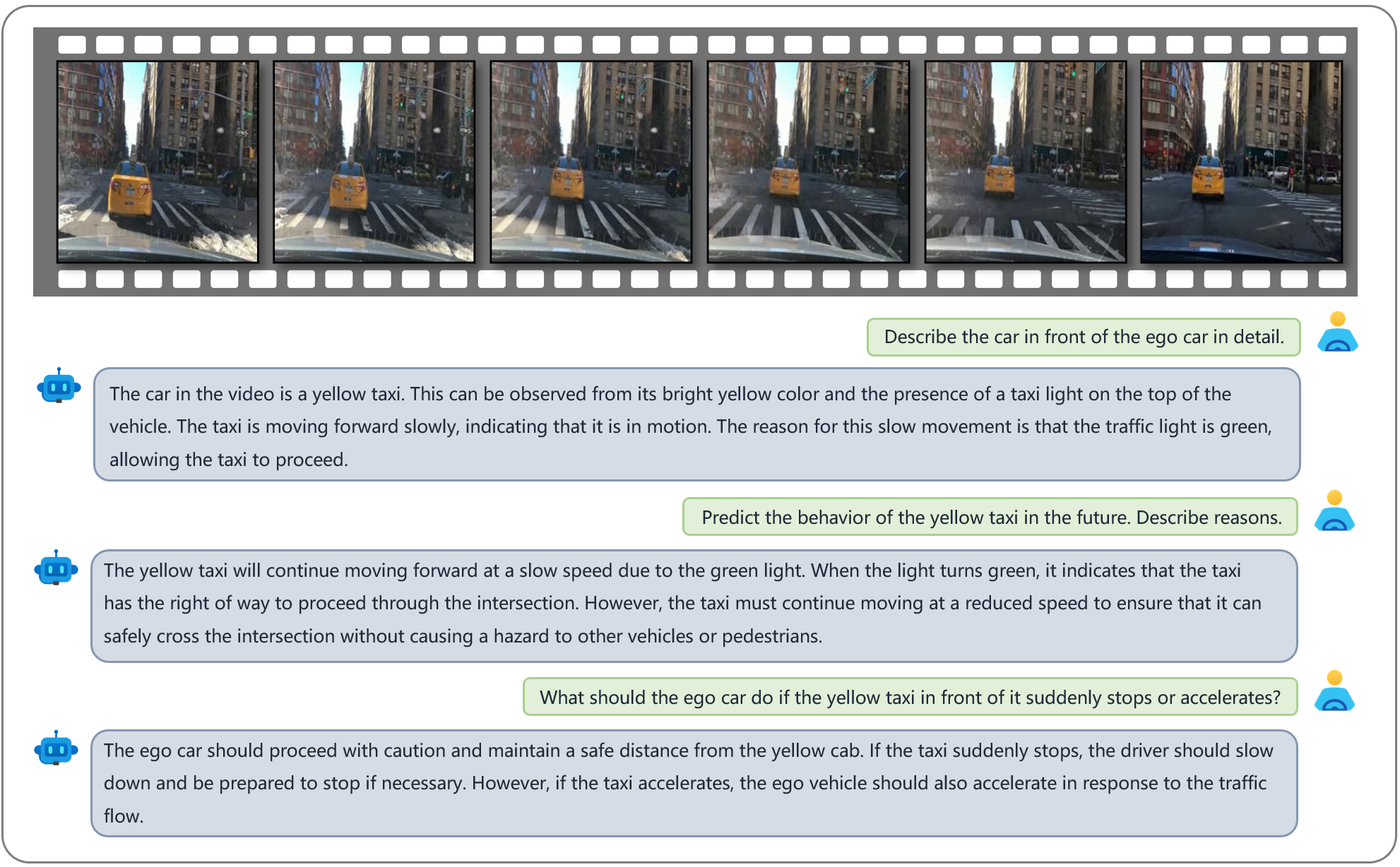}
        \vspace{-2mm}
	\caption{An example showcasing \model’s capability in \textbf{prediction} and \textbf{contingency planning}~\S~\ref{sec:prediction_planning}. The video shows an ego car following a taxi and going through an intersection. On one hand, \model{} can predict the future behavior of the yellow taxi for a certain period. On the other hand, \model{} can make reasonable contingency plans for the ego car in case the yellow taxi in front suddenly accelerates or comes to a stop.}
\label{fig:planning_4}
\end{figure}

Following DriveLM~\cite{drivelm2023}, We also evaluate \model{}'s prediction and planning capabilities, which involve utilizing the \textbf{reasoning} ability of VLMs to assist the driver in making decisions and ensuring explainable planning. In Figures \ref{fig:prediction_1}, \ref{fig:prediction_2}, \ref{fig:planning_4} and \ref{fig:planning_5}, we showcase our model's multimodal ability to predict the behavior of other vehicles in the future and determine whether these vehicles affect the ego agent's trajectory.  Figures \ref{fig:planning_1}, \ref{fig:planning_2}, \ref{fig:planning_3}, and \ref{fig:planning_6} present some examples that demonstrate \model{} can generate comprehensive plans for the ego car based on current traffic conditions. Furthermore, Figures \ref{fig:planning_4} and \ref{fig:planning_5} demonstrate \model{}'s capabilities to make reasonable and safe plans based on the contingent behavior of other agents, which is considered crucial in real-world driving scenarios.
However, due to the lack of relevant instructions during training, we currently recommend using in-context learning ability, ongoing dialogue, and control signals to assist \model{} in completing these two tasks. We believe this is still an under-explored facet and we are working on it.  

\subsection{Human-like Capabilities}
In this set of demonstrations, we will show the human-like capabilities of \model{} in (1) rapid learning and adaptation through in-context learning (\S~\ref{sec:in-context}); (2) error recovering through reflection (\S~\ref{sec:reflection}); and (3) interpretability through interactive conversation (\S~\ref{sec:conversation}).

\subsubsection{Rapid Learning and Adaptation} \label{sec:in-context}
\begin{figure}[!htb]
	\centering
	\includegraphics[width=1.0\textwidth,trim=0 0 0 0,clip]{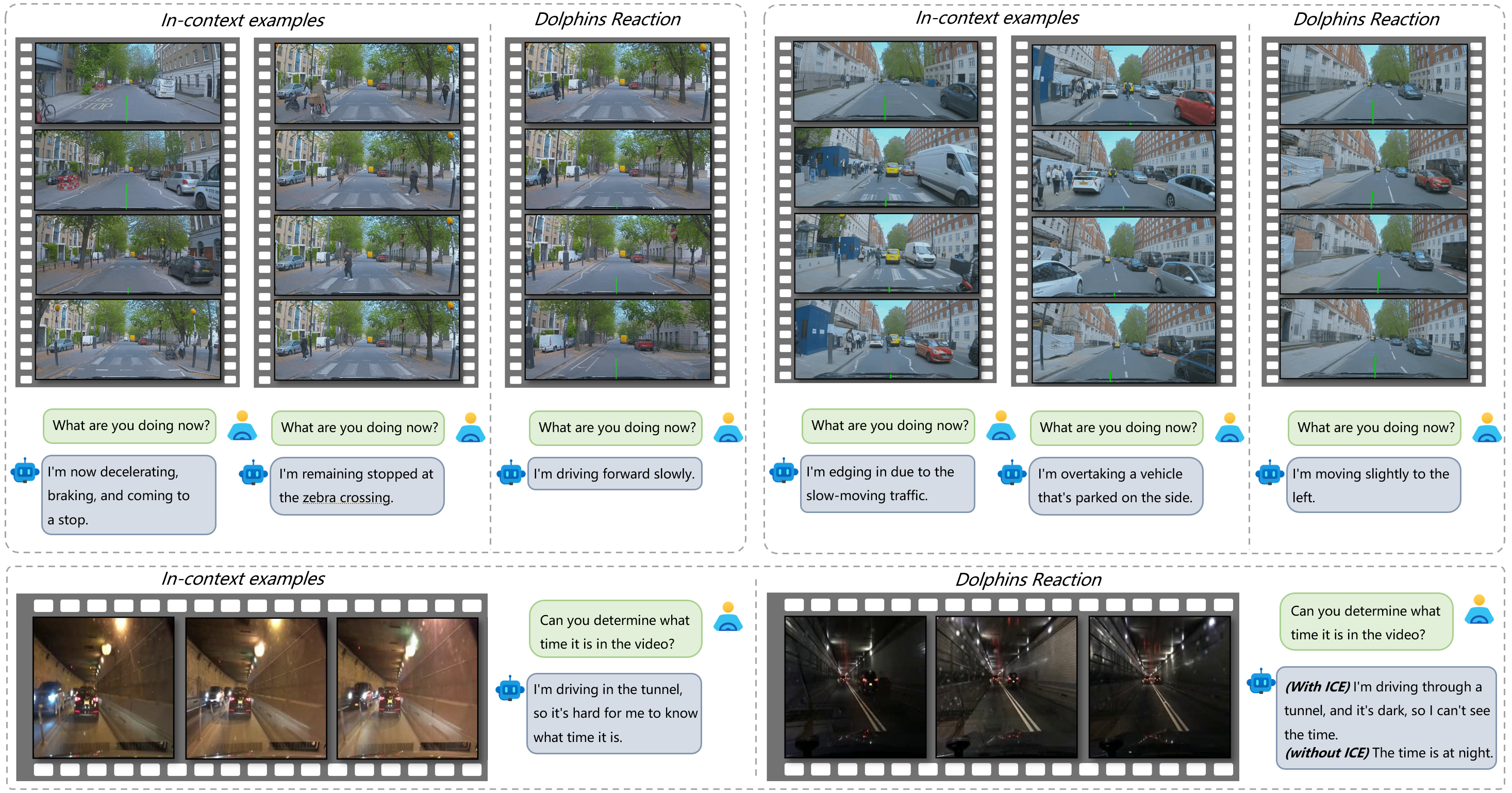}
        \vspace{-2mm}
	\caption{Three examples show our model enables rapid adaptation to unseen instructions through \textbf{in-context learning}~\S~\ref{sec:in-context}. In the first two examples, \model{} learns to play the role of the driver through in-context examples and can accurately describe its behavior, despite not having been trained on such instructions. The third example shows that \model{} can learn common sense from in-context examples, such as not being able to judge the current time based on the light when inside a tunnel.}
\label{fig:in_context_learning_1}
\end{figure}

\begin{figure}[!htb]
	\centering
	\includegraphics[width=1.0\textwidth,trim=0 0 0 0,clip]{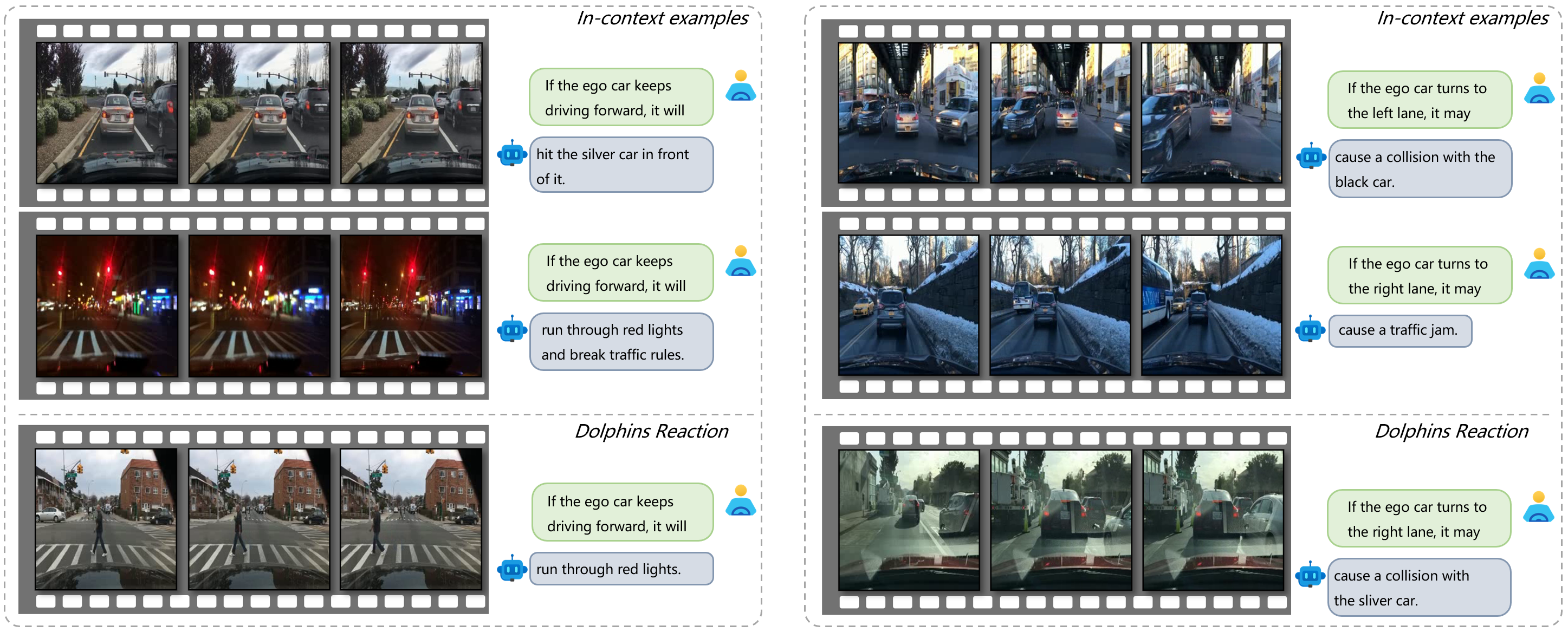}
        \vspace{-2mm}
	\caption{An example showcasing \model’s capability in \textbf{prediction} through \textbf{in-context learning}~\S~\ref{sec:in-context}. \model{} can reason for future events that have not yet happened. Through in-context examples, our model can learn to answer "What If"-style questions, to imagine the behavior of the ego car in the future by language.}
\label{fig:in_context_learning_2}
\end{figure}

In this demonstration, we document the agility of \model{} to rapidly assimilate and adapt to new driving conditions—a process akin to human learning. This facet is examined by presenting \model{} with a series of unforeseen scenarios and monitoring its response efficiency and accuracy after exposure to a limited set of examples. The tasks are designed to test the model's in-context learning capabilities by progressively introducing more complex and previously unseen driving scenarios. Through this, \model{} demonstrates its ability to leverage prior knowledge and quickly adapt, suggesting that even in the absence of extensive pre-training on certain tasks, it can still formulate accurate predictions and actions using a sparse sampling of in-context examples. Specifically, Figure \ref{fig:in_context_learning_1} shows our model's ability to learn certain common-sense knowledge from in-context examples, such as ``\textit{You cannot determine the current time by the light in a tunnel.}''. Besides, \model{} demonstrates the excellent ability to respond to unseen instructions in various styles, such as ``\textit{What if}'' and ``\textit{What are you doing}'', as shown in Figures \ref{fig:in_context_learning_1} and \ref{fig:in_context_learning_2}.

\subsubsection{Interactive Conversation} \label{sec:conversation}
\begin{figure}[!htb]
	\centering
	\includegraphics[width=1.0\textwidth,trim=0 0 0 0,clip]{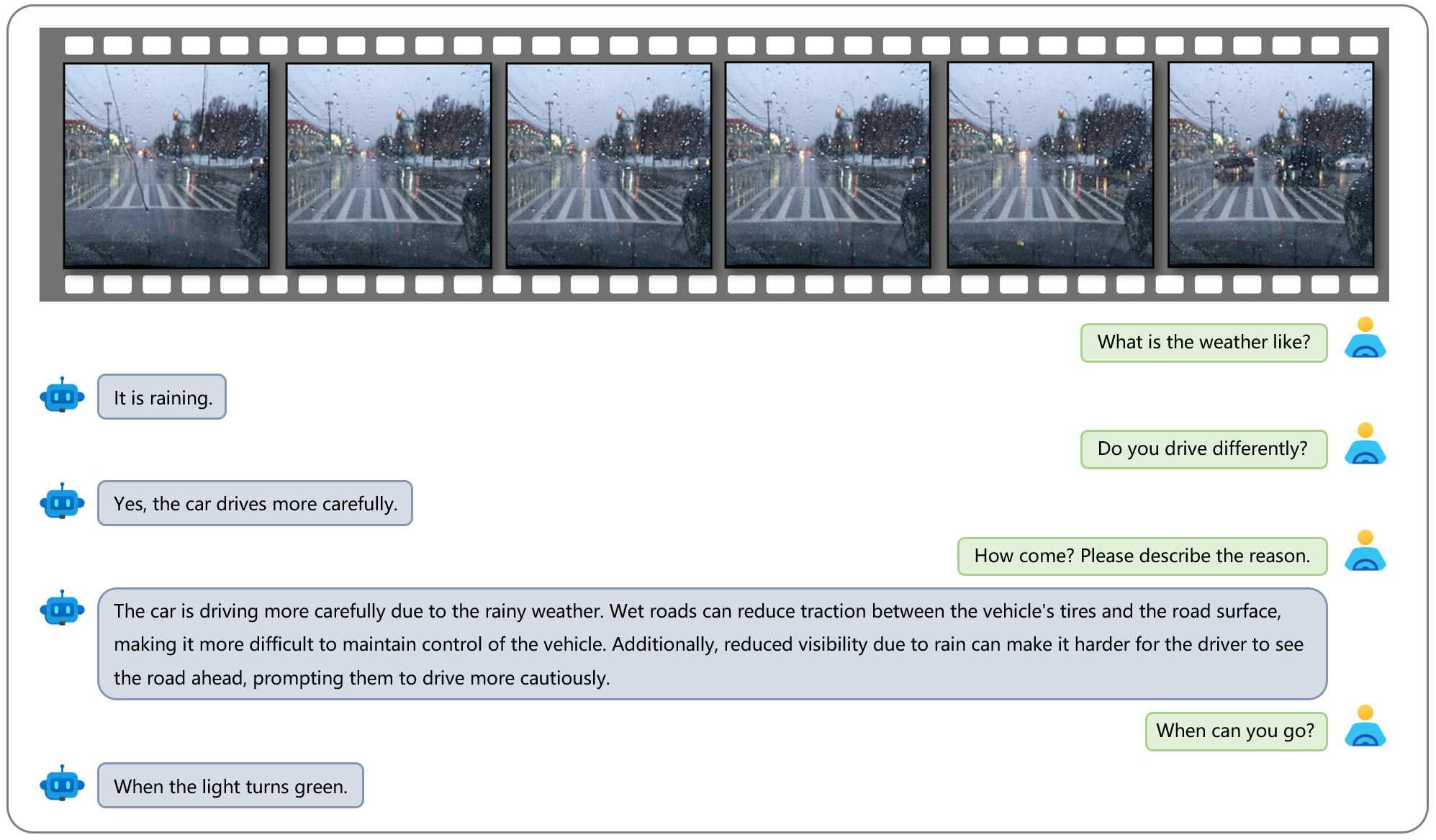}
	\caption{An example showcasing \model’s capability in \textbf{interactive conversation}~\S~\ref{sec:conversation}. The video features an ego car stopping at an intersection on a rainy day, where \model{} can answer various questions about the weather, the traffic lights, and their impact on driving.}
\label{fig:conversation_1}
\end{figure}

\begin{figure}[!htb]
	\centering
 
	\includegraphics[width=1.0\textwidth,trim=0 0 0 0,clip]{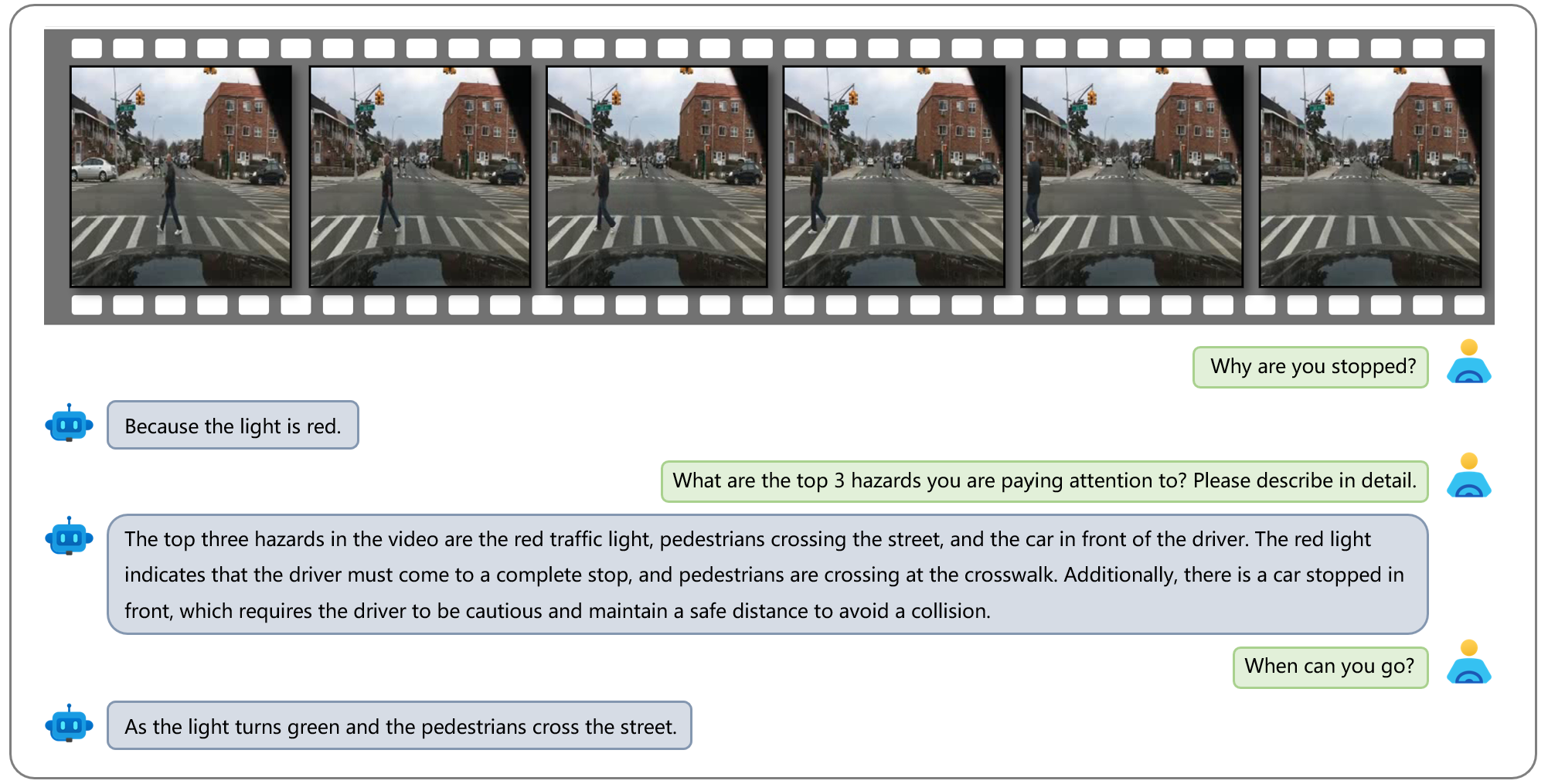}
	\caption{An example showcasing \model’s capability in \textbf{ interactive conversation}~\S~\ref{sec:conversation}. This video shows an ego car stopped at an intersection waiting for a red light and a pedestrian crossing a zebra crossing. We can ask \model{} some questions about what it’s paying attention to at this intersection.}
\label{fig:conversation_2}
\end{figure}

\begin{figure}[!htb]
	\centering
 
	\includegraphics[width=1.0\textwidth,trim=0 0 0 0,clip]{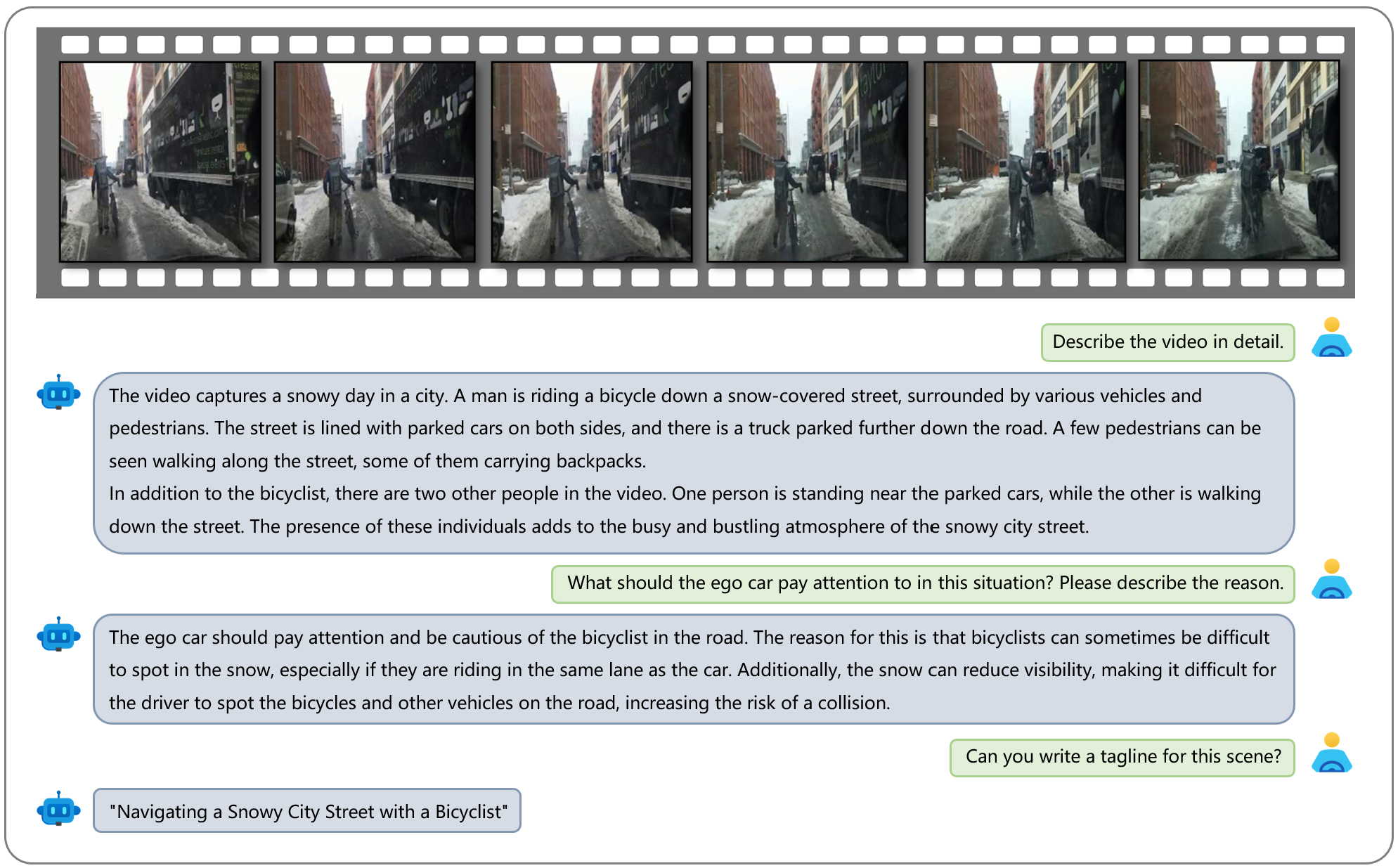}
	\caption{An example showcasing \model’s capability in \textbf{interactive conversation}~\S~\ref{sec:conversation}. This video shows an ego car driving behind a bicyclist. Through conversation with \model, we can understand how to drive safely in the presense of cyclists. And our model can also create a tagline for this scene.}
\label{fig:conversation_4}
\end{figure}

In this demonstration, we subject \model{} to an evaluation of its conversational skills through multi-turn dialogues, gauging its competence in engaging with human drivers under varying conditions. Utilizing a set of instructions primarily derived from LINGO-1~\footnote{https://wayve.ai/thinking/lingo-natural-language-autonomous-driving/}, we present \model{} with a spectrum of queries reflective of real-world driving interactions. As shown in Figures \ref{fig:conversation_1}, \ref{fig:conversation_2}, and \ref{fig:conversation_4}, the conversations are constructed to assess \model{}'s ability to comprehend and respond to nuanced language, maintain context over multiple exchanges, and offer informative and contextually relevant responses spanning from potential hazards in the scene to ego planning and the reasoning behind the scene. The results from these interactions indicate that \model{} possesses a robust conversational ability, distinguishing itself significantly from other contemporary driving-related Vision Language Models in terms of linguistic flexibility and contextual understanding. In the future, this could be a foundation for a human interface that builds up trust between AV and road users or its passengers.

\subsubsection{Reflection and Error Recovering} \label{sec:reflection}
\begin{figure}[!htb]
	\centering
	\includegraphics[width=0.85\textwidth,trim=0 0 0 0,clip]{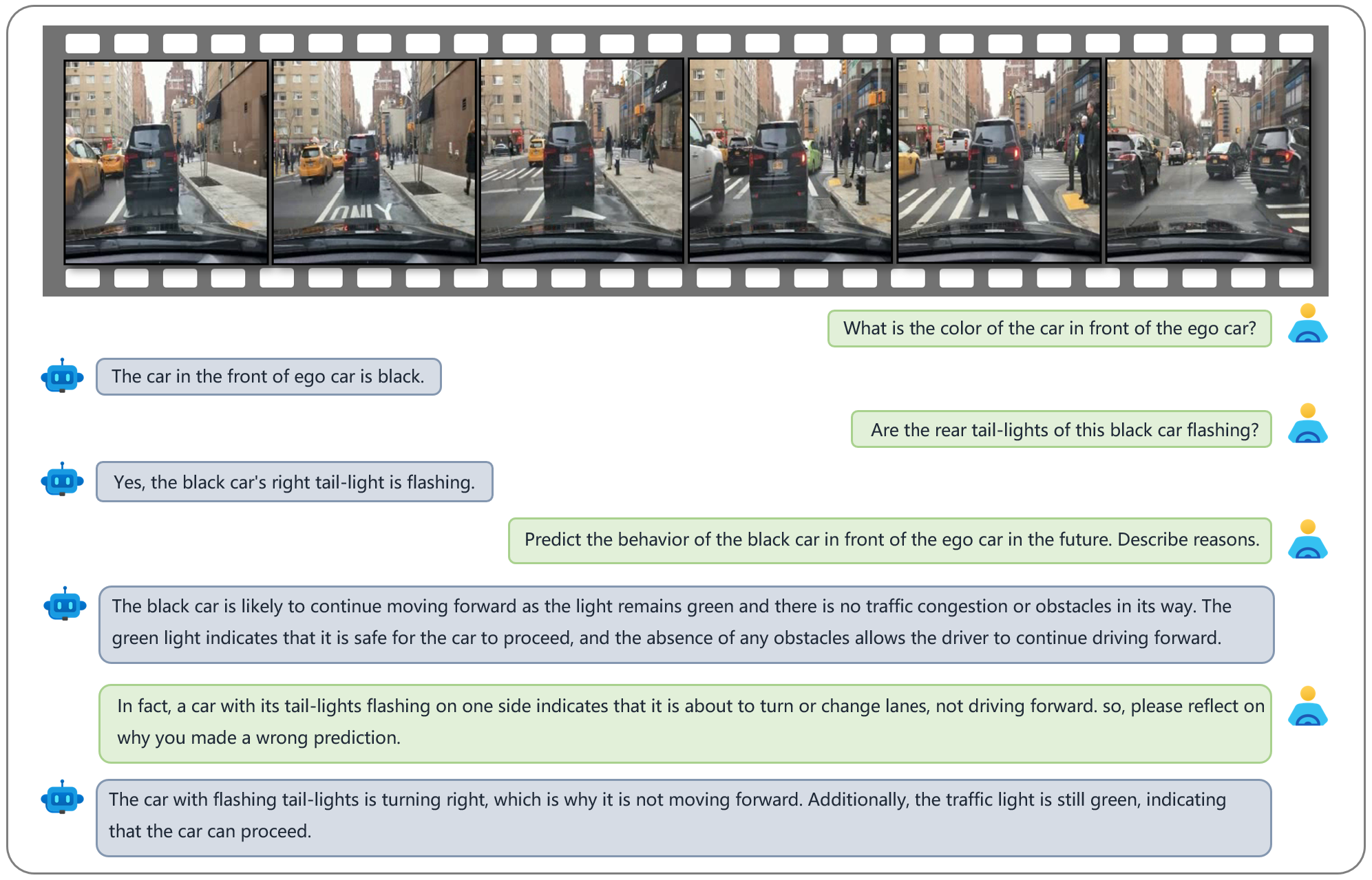}
        \vspace{-2mm}
	\caption{An example showcasing \model’s capability in \textbf{reflection}~\S~\ref{sec:reflection}. The video features an ego car driving through an interaction on a city street. The black vehicle in front of the ego car appears to want to turn right as its right taillight is flashing. Initially, \model{} predicts that the black car will continue to drive forward. However, after being provided with information about the 'slashing rear lights indicate a turn or lane change.' traffic rule, \model{} reflects and revises the wrong prediction.}
\label{fig:reflection_1}
\end{figure}

\begin{figure}[!htb]
	\centering
	\includegraphics[width=0.85\textwidth,trim=0 0 0 0,clip]{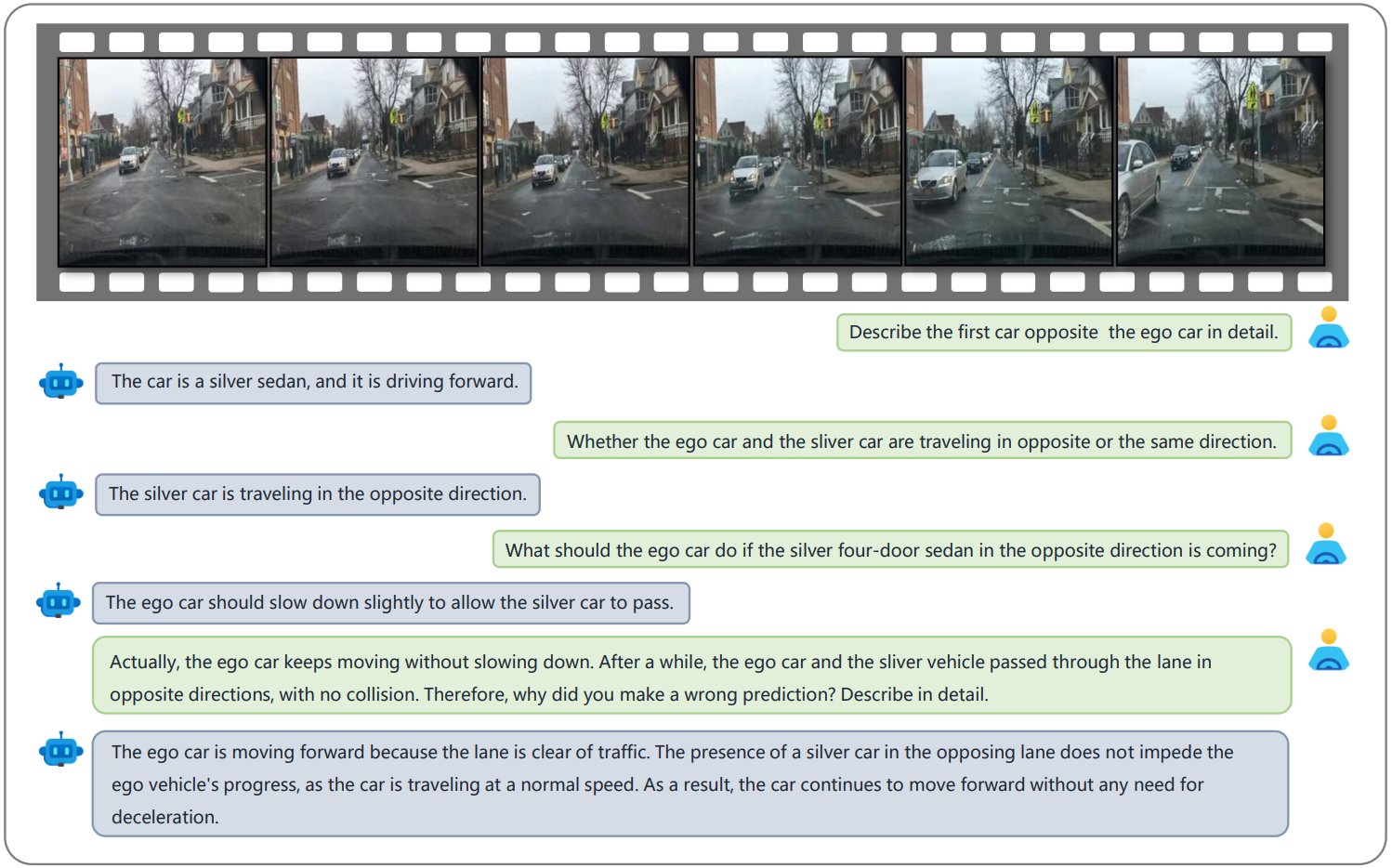}
    \vspace{-2mm}
	\caption{An example showcasing \model’s capability in \textbf{reflection}~\S~\ref{sec:reflection}. The video shows an ego car driving through an interaction with a white car coming from the opposite direction. Initially, \model{} plans for the ego car to slow down and let the white car proceed. However, after being provided with information about the human driver decision, \model{} reflects and revises the driving plan.}
\label{fig:reflection_2}
\end{figure}

\begin{figure}[!htb]
	\centering
	\includegraphics[width=0.85\textwidth,trim=0 0 0 0,clip]{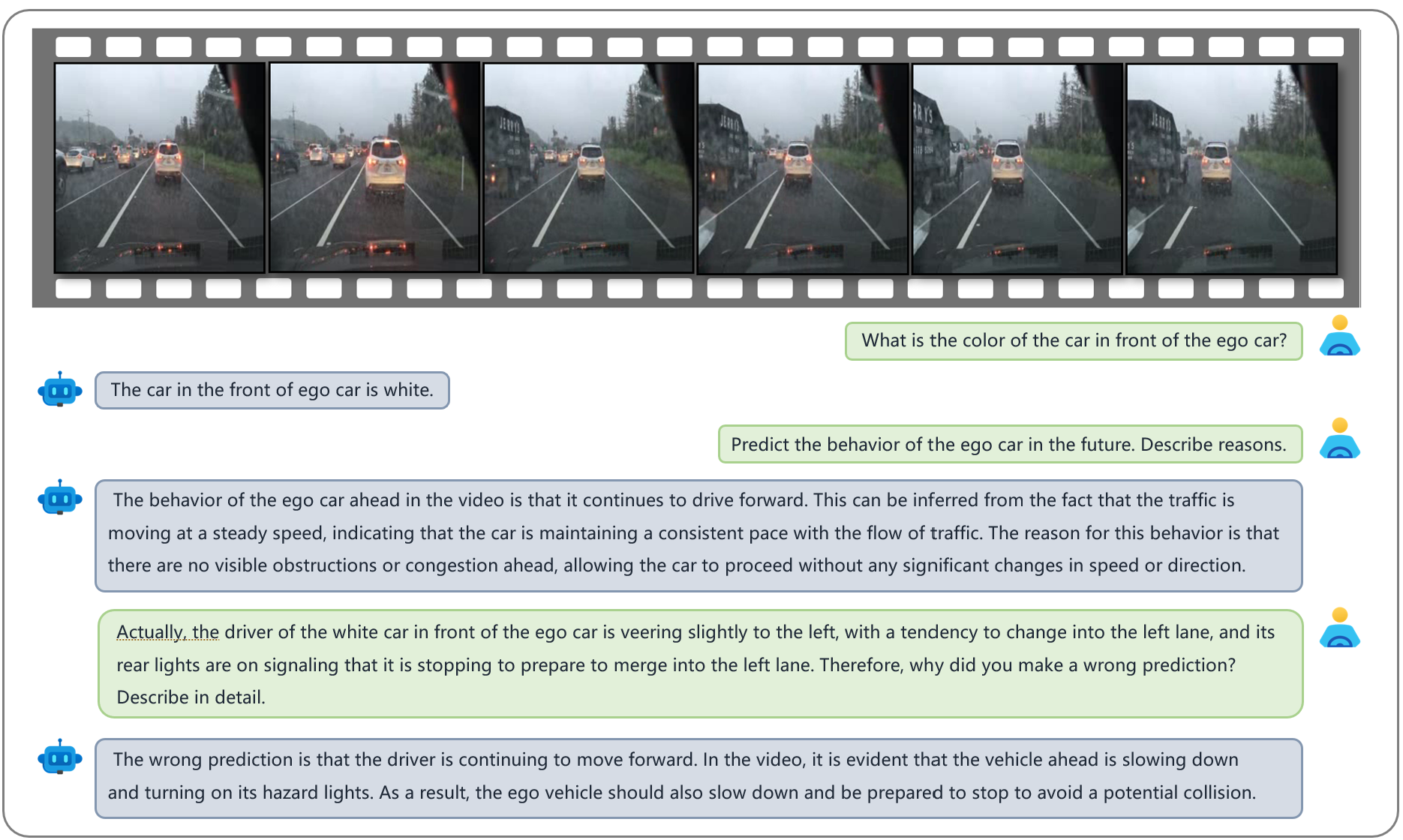}
 \vspace{-2mm}
	\caption{An example showcasing \model’s capability in \textbf{reflection}~\S~\ref{sec:reflection}. The video shows an ego car driving on a highway. Initially, \model{} predicts that the white car in front of the ego car will continue to drive forward. However, after being provided with information about the human driver decision, \model{} revises the wrong prediction and realizes that the ego car should have also stopped to avoid the white car that is about to stop and merge into the left lane.}
\label{fig:reflection_3}
\end{figure}

This demonstration is devoted to showcasing \model{}'s self-assessment and error correction mechanisms. We present instances where the model first produces a suboptimal response to a given driving scenario and is subsequently provided with feedback. The focus here is on how \model{} reflects on this feedback to identify and correct its mistakes. We evaluate the effectiveness of these mechanisms through a series of iterative interactions, illustrating the model's capacity for reflection, error identification, and the implementation of corrective measures. The results (evidenced in Figure~\ref{fig:reflection_1}, \ref{fig:reflection_2} and \ref{fig:reflection_3}) underscore the model's ability to not just detect and recover from errors in a manner that mirrors human cognitive processes, but to also refine its subsequent responses, thereby enhancing overall performance and reliability. For example, Figure~\ref{fig:reflection_1} presents that \model{} is able to reflect on the tail-light signals and predict the correct right turning behavior of the black car; and Figure~\ref{fig:reflection_2} shows that \model{} can reflect and revise the reasoning for the unnecessary yielding decision when encountering a car in a narrow street.

\section{Conclusion and Future Directions}

As we conclude our exploration into \model{}, a novel vision-language model designed for enhancing autonomous vehicles (AVs), we reflect on the significant strides made and the challenges ahead. \model{} has demonstrated a remarkable capacity for holistic understanding and human-like reasoning in complex driving scenarios, marking a substantial advancement in the realm of autonomous driving technology. By leveraging multimodal inputs and employing the innovative Grounded Chain of Thought (GCoT) process, \model{} has shown its proficiency as a conversational driving assistant, capable of addressing a wide spectrum of AV tasks with enhanced interpretability and rapid adaptation capabilities.

However, our journey towards fully optimizing \model{} for real-world application in AVs encounters notable challenges, particularly in computational overhead and feasibility. Our assessment of \model{}'s performance on the DriveLM dataset, a realistic benchmark for real-world driving scenarios, revealed an average inference time of 1.34 seconds on an NVIDIA A100, indicating a potential limitation in achieving high frame rates on edge devices. Additionally, the power consumption associated with running such sophisticated models in vehicles presents a significant hurdle for deployment. These findings underscore the necessity of further advancements in model efficiency.
Looking forward, the development of customized and distilled versions of these models, as suggested by emerging research ~\cite{tiny_chat}, appears to be a promising direction. These streamlined models are anticipated to be more feasible for deployment on edge devices, balancing computational demands with power efficiency. We believe that continued exploration and innovation in this domain are vital for realizing the full potential of AVs equipped with advanced AI capabilities like those offered by \model{}.


  
\begin{figure}[!htb]
	\centering
	\includegraphics[width=1.0\textwidth,trim=0 0 0 0,clip]{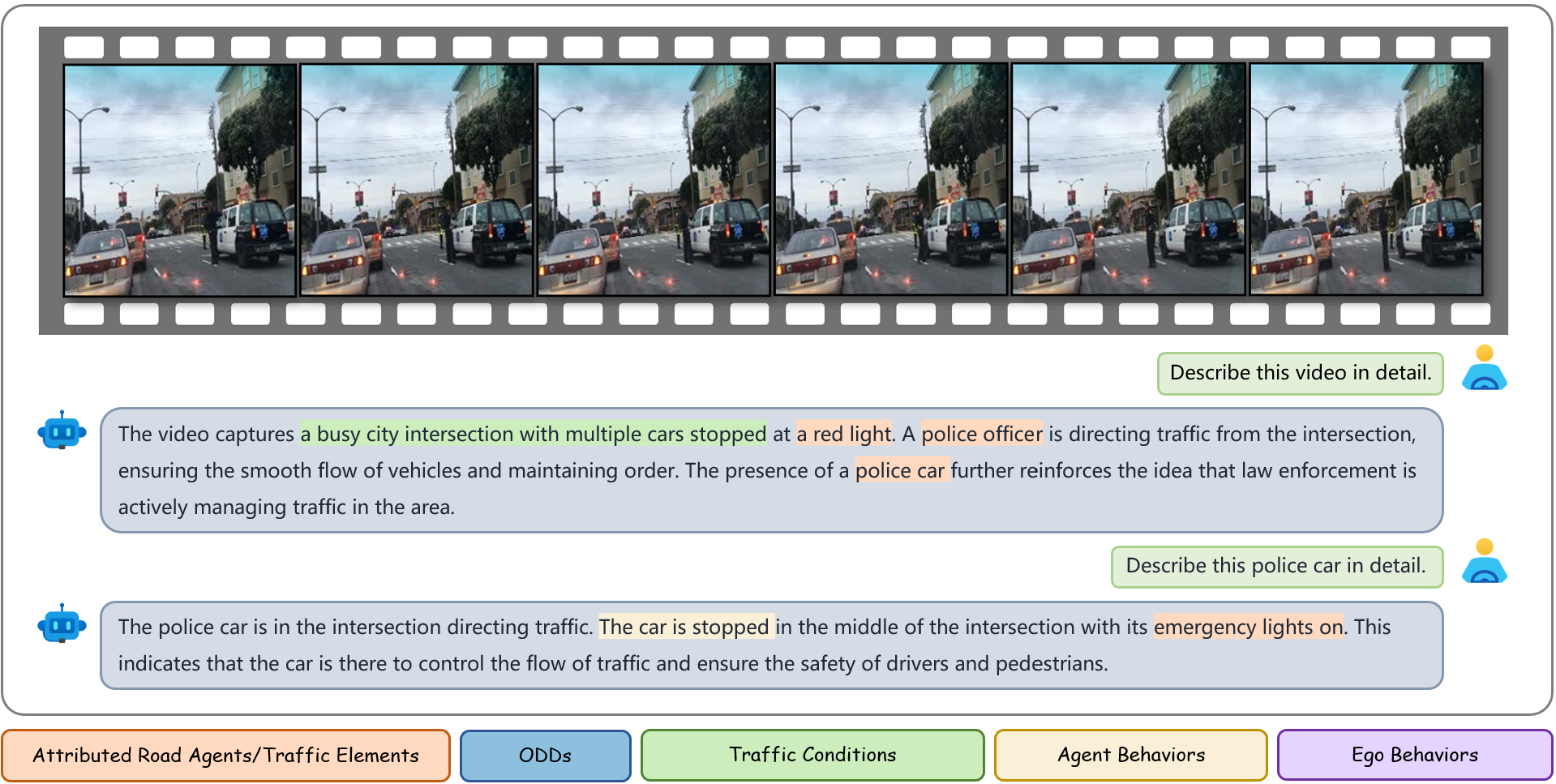}
        \vspace{-2mm}
	\caption{An example showcasing \model’s capability in \textbf{scenario understanding}~\S~\ref{sec:perception}. The video shows an ego car stopped at a busy interaction with a police car next to it. \model{} can identify the police officer and police car with its emergency lights on. }
\label{fig:perception_7}
\end{figure}

\begin{figure}[!htb]
	\centering
	\includegraphics[width=0.85\textwidth,trim=0 0 0 0,clip]{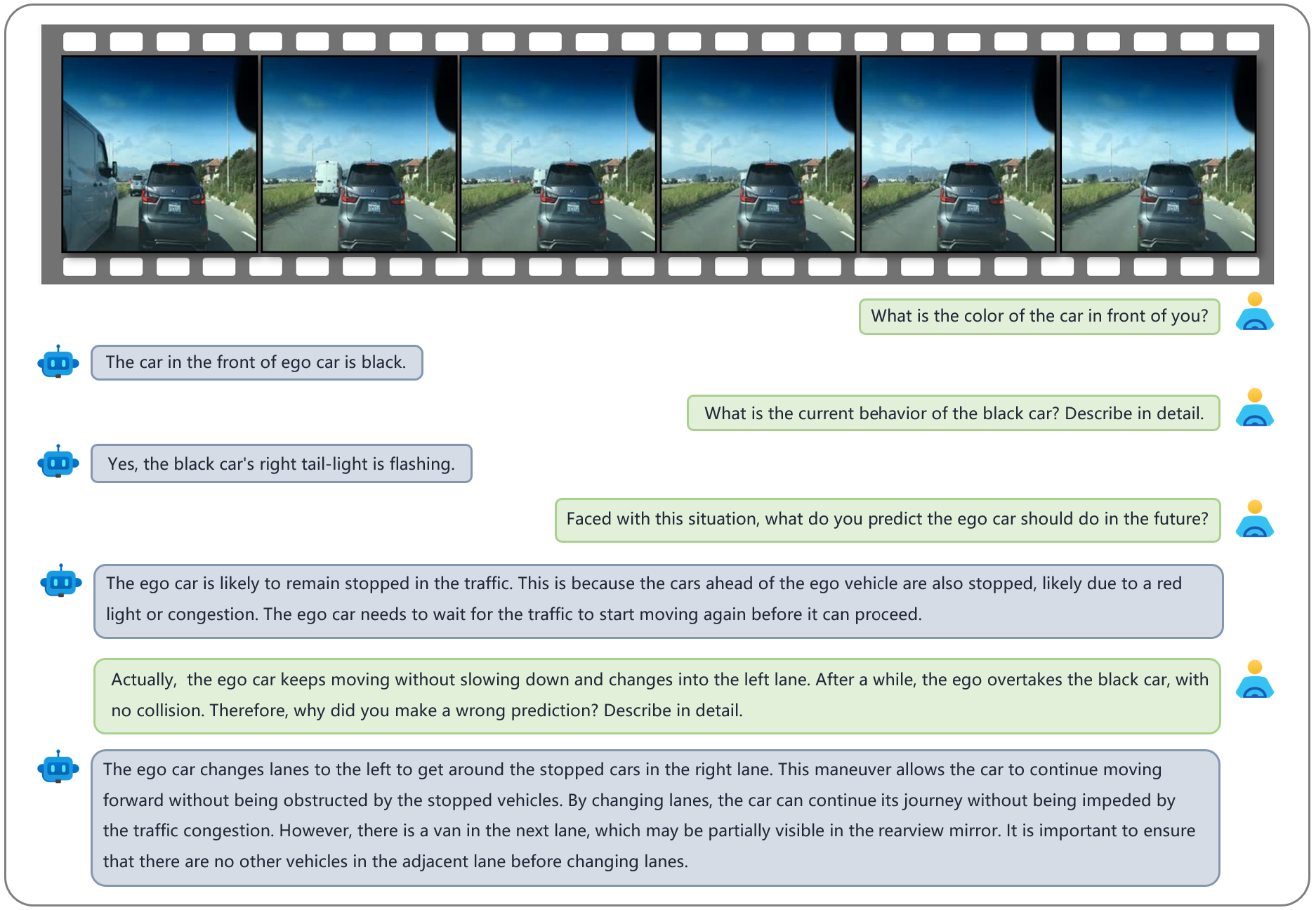}
 \vspace{-2mm}
	\caption{An example showcasing \model’s capability in \textbf{reflection}~\S~\ref{sec:reflection}. The video shows an ego car stopping behind a black car. Initially, \model{} plans for the ego car to remain stopped and wait for the traffic. However, after being provided with information about the human driver decision~(driving more aggressively), \model{} reflects and plans for the ego car to change lanes to the left to continue moving forward when careful attention is paid to cars traveling the left lane, such as a white van.}
\label{fig:reflection_4}
\end{figure}

\clearpage
\bibliographystyle{unsrt}  
\bibliography{references}  

\begin{thebibliography}{10}

\bibitem{jordan2015machine}
Michael~I Jordan and Tom~M Mitchell.
\newblock Machine learning: Trends, perspectives, and prospects.
\newblock {\em Science}, 349(6245):255--260, 2015.

\bibitem{soori2023artificial}
Mohsen Soori, Behrooz Arezoo, and Roza Dastres.
\newblock Artificial intelligence, machine learning and deep learning in
  advanced robotics, a review.
\newblock {\em Cognitive Robotics}, 2023.

\bibitem{mom2023evolution}
Gijs Mom.
\newblock {\em The evolution of automotive technology: a handbook}.
\newblock SAE International, 2023.

\bibitem{li2019aads}
Wei Li, CW~Pan, Rong Zhang, JP~Ren, YX~Ma, Jin Fang, FL~Yan, QC~Geng, XY~Huang,
  HJ~Gong, et~al.
\newblock Aads: Augmented autonomous driving simulation using data-driven
  algorithms.
\newblock {\em Science robotics}, 4(28):eaaw0863, 2019.

\bibitem{tampuu2020survey}
Ardi Tampuu, Tambet Matiisen, Maksym Semikin, Dmytro Fishman, and Naveed
  Muhammad.
\newblock A survey of end-to-end driving: Architectures and training methods.
\newblock {\em IEEE Transactions on Neural Networks and Learning Systems},
  33(4):1364--1384, 2020.

\bibitem{coelho2022review}
Daniel Coelho and Miguel Oliveira.
\newblock A review of end-to-end autonomous driving in urban environments.
\newblock {\em IEEE Access}, 10:75296--75311, 2022.

\bibitem{chen2023end}
Li~Chen, Penghao Wu, Kashyap Chitta, Bernhard Jaeger, Andreas Geiger, and
  Hongyang Li.
\newblock End-to-end autonomous driving: Challenges and frontiers.
\newblock {\em arXiv preprint arXiv:2306.16927}, 2023.

\bibitem{liu2020computing}
Liangkai Liu, Sidi Lu, Ren Zhong, Baofu Wu, Yongtao Yao, Qingyang Zhang, and
  Weisong Shi.
\newblock Computing systems for autonomous driving: State of the art and
  challenges.
\newblock {\em IEEE Internet of Things Journal}, 8(8):6469--6486, 2020.

\bibitem{jain2021autonomy}
Ashesh Jain, Luca Del~Pero, Hugo Grimmett, and Peter Ondruska.
\newblock Autonomy 2.0: Why is self-driving always 5 years away?
\newblock {\em arXiv preprint arXiv:2107.08142}, 2021.

\bibitem{wong2020mapping}
Kelvin Wong, Yanlei Gu, and Shunsuke Kamijo.
\newblock Mapping for autonomous driving: Opportunities and challenges.
\newblock {\em IEEE Intelligent Transportation Systems Magazine},
  13(1):91--106, 2020.

\bibitem{codevilla2019exploring}
Felipe Codevilla, Eder Santana, Antonio~M L{\'o}pez, and Adrien Gaidon.
\newblock Exploring the limitations of behavior cloning for autonomous driving.
\newblock In {\em Proceedings of the IEEE/CVF International Conference on
  Computer Vision}, pages 9329--9338, 2019.

\bibitem{de2020evaluating}
Lucas de~Paula~Veronese, Fernando Auat-Cheein, Filipe Mutz, Thiago
  Oliveira-Santos, Jos{\'e}~E Guivant, Edilson De~Aguiar, Claudine Badue, and
  Alberto~Ferreira De~Souza.
\newblock Evaluating the limits of a lidar for an autonomous driving
  localization.
\newblock {\em IEEE Transactions on Intelligent Transportation Systems},
  22(3):1449--1458, 2020.

\bibitem{levinson2011towards}
Jesse Levinson, Jake Askeland, Jan Becker, Jennifer Dolson, David Held, Soeren
  Kammel, J~Zico Kolter, Dirk Langer, Oliver Pink, Vaughan Pratt, et~al.
\newblock Towards fully autonomous driving: Systems and algorithms.
\newblock In {\em 2011 IEEE intelligent vehicles symposium (IV)}, pages
  163--168. IEEE, 2011.

\bibitem{yurtsever2020survey}
Ekim Yurtsever, Jacob Lambert, Alexander Carballo, and Kazuya Takeda.
\newblock A survey of autonomous driving: Common practices and emerging
  technologies.
\newblock {\em IEEE access}, 8:58443--58469, 2020.

\bibitem{Liu2023VisualIT}
Haotian Liu, Chunyuan Li, Qingyang Wu, and Yong~Jae Lee.
\newblock Visual instruction tuning.
\newblock {\em ArXiv}, abs/2304.08485, 2023.

\bibitem{touvron2023llama1}
Hugo Touvron, Thibaut Lavril, Gautier Izacard, Xavier Martinet, Marie-Anne
  Lachaux, Timoth{\'e}e Lacroix, Baptiste Rozi{\`e}re, Naman Goyal, Eric
  Hambro, Faisal Azhar, et~al.
\newblock Llama: Open and efficient foundation language models.
\newblock {\em arXiv preprint arXiv:2302.13971}, 2023.

\bibitem{li2023llava}
Chunyuan Li, Cliff Wong, Sheng Zhang, Naoto Usuyama, Haotian Liu, Jianwei Yang,
  Tristan Naumann, Hoifung Poon, and Jianfeng Gao.
\newblock Llava-med: Training a large language-and-vision assistant for
  biomedicine in one day.
\newblock {\em arXiv preprint arXiv:2306.00890}, 2023.

\bibitem{Awadalla2023OpenFlamingoAO}
Anas Awadalla, Irena Gao, Josh Gardner, Jack Hessel, Yusuf Hanafy, Wanrong Zhu,
  Kalyani Marathe, Yonatan Bitton, Samir~Yitzhak Gadre, Shiori Sagawa, Jenia
  Jitsev, Simon Kornblith, Pang~Wei Koh, Gabriel Ilharco, Mitchell Wortsman,
  and Ludwig Schmidt.
\newblock Openflamingo: An open-source framework for training large
  autoregressive vision-language models.
\newblock {\em ArXiv}, abs/2308.01390, 2023.

\bibitem{goyal2017making}
Yash Goyal, Tejas Khot, Douglas Summers-Stay, Dhruv Batra, and Devi Parikh.
\newblock Making the v in vqa matter: Elevating the role of image understanding
  in visual question answering.
\newblock In {\em Proceedings of the IEEE conference on computer vision and
  pattern recognition}, pages 6904--6913, 2017.

\bibitem{marino2019ok}
Kenneth Marino, Mohammad Rastegari, Ali Farhadi, and Roozbeh Mottaghi.
\newblock Ok-vqa: A visual question answering benchmark requiring external
  knowledge.
\newblock In {\em Proceedings of the IEEE/cvf conference on computer vision and
  pattern recognition}, pages 3195--3204, 2019.

\bibitem{hudson2019gqa}
Drew~A Hudson and Christopher~D Manning.
\newblock Gqa: A new dataset for real-world visual reasoning and compositional
  question answering.
\newblock In {\em Proceedings of the IEEE/CVF conference on computer vision and
  pattern recognition}, pages 6700--6709, 2019.

\bibitem{kafle2017analysis}
Kushal Kafle and Christopher Kanan.
\newblock An analysis of visual question answering algorithms.
\newblock In {\em Proceedings of the IEEE international conference on computer
  vision}, pages 1965--1973, 2017.

\bibitem{zhao2023svit}
Bo~Zhao, Boya Wu, and Tiejun Huang.
\newblock Svit: Scaling up visual instruction tuning.
\newblock {\em arXiv preprint arXiv:2307.04087}, 2023.

\bibitem{Kim2018TextualEF}
Jinkyu Kim, Anna Rohrbach, Trevor Darrell, John~F. Canny, and Zeynep Akata.
\newblock Textual explanations for self-driving vehicles.
\newblock In {\em European Conference on Computer Vision}, 2018.

\bibitem{fu2023drive}
Daocheng Fu, Xin Li, Licheng Wen, Min Dou, Pinlong Cai, Botian Shi, and
  Yu~Qiao.
\newblock Drive like a human: Rethinking autonomous driving with large language
  models.
\newblock {\em arXiv preprint arXiv:2307.07162}, 2023.

\bibitem{Mao2023GPTDriverLT}
Jiageng Mao, Yuxi Qian, Hang Zhao, and Yue Wang.
\newblock Gpt-driver: Learning to drive with gpt.
\newblock {\em ArXiv}, abs/2310.01415, 2023.

\bibitem{jin2023surrealdriver}
Ye~Jin, Xiaoxi Shen, Huiling Peng, Xiaoan Liu, Jingli Qin, Jiayang Li, Jintao
  Xie, Peizhong Gao, Guyue Zhou, and Jiangtao Gong.
\newblock Surrealdriver: Designing generative driver agent simulation framework
  in urban contexts based on large language model.
\newblock {\em arXiv preprint arXiv:2309.13193}, 2023.

\bibitem{drivelm2023}
DriveLM Contributors.
\newblock Drivelm: Drive on language.
\newblock \url{https://github.com/OpenDriveLab/DriveLM}, 2023.

\bibitem{Wu2023LanguagePF}
Dongming Wu, Wencheng Han, Tiancai Wang, Ying-Hao Liu, Xiangyu Zhang, and
  Jianbing Shen.
\newblock Language prompt for autonomous driving.
\newblock {\em ArXiv}, abs/2309.04379, 2023.

\bibitem{nuscenes2019}
Holger Caesar, Varun Bankiti, Alex~H. Lang, Sourabh Vora, Venice~Erin Liong,
  Qiang Xu, Anush Krishnan, Yu~Pan, Giancarlo Baldan, and Oscar Beijbom.
\newblock nuscenes: A multimodal dataset for autonomous driving.
\newblock {\em arXiv preprint arXiv:1903.11027}, 2019.

\bibitem{Touvron2023LLaMAOA}
Hugo Touvron, Thibaut Lavril, Gautier Izacard, Xavier Martinet, Marie-Anne
  Lachaux, Timoth{\'e}e Lacroix, Baptiste Rozi{\`e}re, Naman Goyal, Eric
  Hambro, Faisal Azhar, Aurelien Rodriguez, Armand Joulin, Edouard Grave, and
  Guillaume Lample.
\newblock Llama: Open and efficient foundation language models.
\newblock {\em ArXiv}, abs/2302.13971, 2023.

\bibitem{Touvron2023Llama2O}
Hugo Touvron, Louis Martin, Kevin~R. Stone, Peter Albert, Amjad Almahairi,
  Yasmine Babaei, Nikolay Bashlykov, Soumya Batra, Prajjwal Bhargava, Shruti
  Bhosale, Daniel~M. Bikel, Lukas Blecher, Cristian~Cant{\'o}n Ferrer, Moya
  Chen, Guillem Cucurull, David Esiobu, Jude Fernandes, Jeremy Fu, Wenyin Fu,
  Brian Fuller, Cynthia Gao, Vedanuj Goswami, Naman Goyal, Anthony~S.
  Hartshorn, Saghar Hosseini, Rui Hou, Hakan Inan, Marcin Kardas, Viktor
  Kerkez, Madian Khabsa, Isabel~M. Kloumann, A.~V. Korenev, Punit~Singh Koura,
  Marie-Anne Lachaux, Thibaut Lavril, Jenya Lee, Diana Liskovich, Yinghai Lu,
  Yuning Mao, Xavier Martinet, Todor Mihaylov, Pushkar Mishra, Igor Molybog,
  Yixin Nie, Andrew Poulton, Jeremy Reizenstein, Rashi Rungta, Kalyan Saladi,
  Alan Schelten, Ruan Silva, Eric~Michael Smith, R.~Subramanian, Xia Tan, Binh
  Tang, Ross Taylor, Adina Williams, Jian~Xiang Kuan, Puxin Xu, Zhengxu Yan,
  Iliyan Zarov, Yuchen Zhang, Angela Fan, Melanie Kambadur, Sharan Narang,
  Aurelien Rodriguez, Robert Stojnic, Sergey Edunov, and Thomas Scialom.
\newblock Llama 2: Open foundation and fine-tuned chat models.
\newblock {\em ArXiv}, abs/2307.09288, 2023.

\bibitem{vicuna2023}
Wei-Lin Chiang, Zhuohan Li, Zi~Lin, Ying Sheng, Zhanghao Wu, Hao Zhang, Lianmin
  Zheng, Siyuan Zhuang, Yonghao Zhuang, Joseph~E. Gonzalez, Ion Stoica, and
  Eric~P. Xing.
\newblock Vicuna: An open-source chatbot impressing gpt-4 with 90\%* chatgpt
  quality, March 2023.

\bibitem{MosaicML2023Introducing}
MosaicML~NLP Team.
\newblock Introducing mpt-7b: A new standard for open-source, commercially
  usable llms, 2023.
\newblock Accessed: 2023-05-05.

\bibitem{Alayrac2022FlamingoAV}
Jean-Baptiste Alayrac, Jeff Donahue, Pauline Luc, Antoine Miech, Iain Barr,
  Yana Hasson, Karel Lenc, Arthur Mensch, Katie Millican, Malcolm Reynolds,
  Roman Ring, Eliza Rutherford, Serkan Cabi, Tengda Han, Zhitao Gong, Sina
  Samangooei, Marianne Monteiro, Jacob Menick, Sebastian Borgeaud, Andy Brock,
  Aida Nematzadeh, Sahand Sharifzadeh, Mikolaj Binkowski, Ricardo Barreira,
  Oriol Vinyals, Andrew Zisserman, and Karen Simonyan.
\newblock Flamingo: a visual language model for few-shot learning.
\newblock {\em ArXiv}, abs/2204.14198, 2022.

\bibitem{Li2023BLIP2BL}
Junnan Li, Dongxu Li, Silvio Savarese, and Steven C.~H. Hoi.
\newblock Blip-2: Bootstrapping language-image pre-training with frozen image
  encoders and large language models.
\newblock {\em ArXiv}, abs/2301.12597, 2023.

\bibitem{Zhu2023MiniGPT4EV}
Deyao Zhu, Jun Chen, Xiaoqian Shen, Xiang Li, and Mohamed Elhoseiny.
\newblock Minigpt-4: Enhancing vision-language understanding with advanced
  large language models.
\newblock {\em ArXiv}, abs/2304.10592, 2023.

\bibitem{Li2023OtterAM}
Bo~Li, Yuanhan Zhang, Liangyu Chen, Jinghao Wang, Jingkang Yang, and Ziwei Liu.
\newblock Otter: A multi-modal model with in-context instruction tuning.
\newblock {\em ArXiv}, abs/2305.03726, 2023.

\bibitem{Dai2023InstructBLIPTG}
Wenliang Dai, Junnan Li, Dongxu Li, Anthony Meng~Huat Tiong, Junqi Zhao,
  Weisheng Wang, Boyang~Albert Li, Pascale Fung, and Steven C.~H. Hoi.
\newblock Instructblip: Towards general-purpose vision-language models with
  instruction tuning.
\newblock {\em ArXiv}, abs/2305.06500, 2023.

\bibitem{Ye2023mPLUGOwlME}
Qinghao Ye, Haiyang Xu, Guohai Xu, Jiabo Ye, Ming Yan, Yi~Zhou, Junyan Wang,
  Anwen Hu, Pengcheng Shi, Yaya Shi, Chenliang Li, Yuanhong Xu, Hehong Chen,
  Junfeng Tian, Qiang Qi, Ji~Zhang, and Feiyan Huang.
\newblock mplug-owl: Modularization empowers large language models with
  multimodality.
\newblock {\em ArXiv}, abs/2304.14178, 2023.

\bibitem{li2023videochat}
KunChang Li, Yinan He, Yi~Wang, Yizhuo Li, Wenhai Wang, Ping Luo, Yali Wang,
  Limin Wang, and Yu~Qiao.
\newblock Videochat: Chat-centric video understanding.
\newblock {\em arXiv preprint arXiv:2305.06355}, 2023.

\bibitem{zhang2023video}
Hang Zhang, Xin Li, and Lidong Bing.
\newblock Video-llama: An instruction-tuned audio-visual language model for
  video understanding.
\newblock {\em arXiv preprint arXiv:2306.02858}, 2023.

\bibitem{maaz2023video}
Muhammad Maaz, Hanoona Rasheed, Salman Khan, and Fahad~Shahbaz Khan.
\newblock Video-chatgpt: Towards detailed video understanding via large vision
  and language models.
\newblock {\em arXiv preprint arXiv:2306.05424}, 2023.

\bibitem{luo2023valley}
Ruipu Luo, Ziwang Zhao, Min Yang, Junwei Dong, Minghui Qiu, Pengcheng Lu, Tao
  Wang, and Zhongyu Wei.
\newblock Valley: Video assistant with large language model enhanced ability.
\newblock {\em arXiv preprint arXiv:2306.07207}, 2023.

\bibitem{driess2023palm}
Danny Driess, Fei Xia, Mehdi~SM Sajjadi, Corey Lynch, Aakanksha Chowdhery,
  Brian Ichter, Ayzaan Wahid, Jonathan Tompson, Quan Vuong, Tianhe Yu, et~al.
\newblock Palm-e: An embodied multimodal language model.
\newblock {\em arXiv preprint arXiv:2303.03378}, 2023.

\bibitem{mu2023embodiedgpt}
Yao Mu, Qinglong Zhang, Mengkang Hu, Wenhai Wang, Mingyu Ding, Jun Jin, Bin
  Wang, Jifeng Dai, Yu~Qiao, and Ping Luo.
\newblock Embodiedgpt: Vision-language pre-training via embodied chain of
  thought.
\newblock {\em arXiv preprint arXiv:2305.15021}, 2023.

\bibitem{brohan2023rt}
Anthony Brohan, Noah Brown, Justice Carbajal, Yevgen Chebotar, Xi~Chen,
  Krzysztof Choromanski, Tianli Ding, Danny Driess, Avinava Dubey, Chelsea
  Finn, et~al.
\newblock Rt-2: Vision-language-action models transfer web knowledge to robotic
  control.
\newblock {\em arXiv preprint arXiv:2307.15818}, 2023.

\bibitem{yang2023octopus}
Jingkang Yang, Yuhao Dong, Shuai Liu, Bo~Li, Ziyue Wang, Chencheng Jiang,
  Haoran Tan, Jiamu Kang, Yuanhan Zhang, Kaiyang Zhou, et~al.
\newblock Octopus: Embodied vision-language programmer from environmental
  feedback.
\newblock {\em arXiv preprint arXiv:2310.08588}, 2023.

\bibitem{hong20233d}
Yining Hong, Haoyu Zhen, Peihao Chen, Shuhong Zheng, Yilun Du, Zhenfang Chen,
  and Chuang Gan.
\newblock 3d-llm: Injecting the 3d world into large language models.
\newblock {\em arXiv preprint arXiv:2307.12981}, 2023.

\bibitem{xu2023pointllm}
Runsen Xu, Xiaolong Wang, Tai Wang, Yilun Chen, Jiangmiao Pang, and Dahua Lin.
\newblock Pointllm: Empowering large language models to understand point
  clouds.
\newblock {\em arXiv preprint arXiv:2308.16911}, 2023.

\bibitem{han2023medalpaca}
Tianyu Han, Lisa~C Adams, Jens-Michalis Papaioannou, Paul Grundmann, Tom
  Oberhauser, Alexander L{\"o}ser, Daniel Truhn, and Keno~K Bressem.
\newblock Medalpaca--an open-source collection of medical conversational ai
  models and training data.
\newblock {\em arXiv preprint arXiv:2304.08247}, 2023.

\bibitem{moor2023med}
Michael Moor, Qian Huang, Shirley Wu, Michihiro Yasunaga, Cyril Zakka, Yash
  Dalmia, Eduardo~Pontes Reis, Pranav Rajpurkar, and Jure Leskovec.
\newblock Med-flamingo: a multimodal medical few-shot learner.
\newblock {\em arXiv preprint arXiv:2307.15189}, 2023.

\bibitem{zheng2023marinegpt}
Ziqiang Zheng, Jipeng Zhang, Tuan-Anh Vu, Shizhe Diao, Yue Him~Wong Tim, and
  Sai-Kit Yeung.
\newblock Marinegpt: Unlocking secrets of ocean to the public.
\newblock {\em arXiv preprint arXiv:2310.13596}, 2023.

\bibitem{ding2023hilm}
Xinpeng Ding, Jianhua Han, Hang Xu, Wei Zhang, and Xiaomeng Li.
\newblock Hilm-d: Towards high-resolution understanding in multimodal large
  language models for autonomous driving.
\newblock {\em arXiv preprint arXiv:2309.05186}, 2023.

\bibitem{Xu2023DriveGPT4IE}
Zhenhua Xu, Yujia Zhang, Enze Xie, Zhen Zhao, Yong Guo, Kenneth~K.Y. Wong,
  Zhenguo Li, and Hengshuang Zhao.
\newblock Drivegpt4: Interpretable end-to-end autonomous driving via large
  language model.
\newblock {\em ArXiv}, abs/2310.01412, 2023.

\bibitem{Li2023MIMICITMI}
Bo~Li, Yuanhan Zhang, Liangyu Chen, Jinghao Wang, Fanyi Pu, Jingkang Yang,
  C.~Li, and Ziwei Liu.
\newblock Mimic-it: Multi-modal in-context instruction tuning.
\newblock {\em ArXiv}, abs/2306.05425, 2023.

\bibitem{Min2021MetaICLLT}
Sewon Min, Mike Lewis, Luke Zettlemoyer, and Hannaneh Hajishirzi.
\newblock Metaicl: Learning to learn in context.
\newblock {\em ArXiv}, abs/2110.15943, 2021.

\bibitem{zhang2023gpt4roi}
Shilong Zhang, Peize Sun, Shoufa Chen, Min Xiao, Wenqi Shao, Wenwei Zhang, Kai
  Chen, and Ping Luo.
\newblock Gpt4roi: Instruction tuning large language model on
  region-of-interest.
\newblock {\em arXiv preprint arXiv:2307.03601}, 2023.

\bibitem{chen2023shikra}
Keqin Chen, Zhao Zhang, Weili Zeng, Richong Zhang, Feng Zhu, and Rui Zhao.
\newblock Shikra: Unleashing multimodal llm's referential dialogue magic.
\newblock {\em arXiv preprint arXiv:2306.15195}, 2023.

\bibitem{rubin2021learning}
Ohad Rubin, Jonathan Herzig, and Jonathan Berant.
\newblock Learning to retrieve prompts for in-context learning.
\newblock {\em arXiv preprint arXiv:2112.08633}, 2021.

\bibitem{openai_chat}
Openai chat.
\newblock \url{https://chat.openai.com}.
\newblock Accessed: 2023-10-20.

\bibitem{Chen2021MetalearningVL}
Yanda Chen, Ruiqi Zhong, Sheng Zha, George Karypis, and He~He.
\newblock Meta-learning via language model in-context tuning.
\newblock {\em ArXiv}, abs/2110.07814, 2021.

\bibitem{Iyer2022OPTIMLSL}
Srinivas Iyer, Xiaojuan Lin, Ramakanth Pasunuru, Todor Mihaylov, Daniel Simig,
  Ping Yu, Kurt Shuster, Tianlu Wang, Qing Liu, Punit~Singh Koura, Xian Li,
  Brian O'Horo, Gabriel Pereyra, Jeff Wang, Christopher Dewan, Asli
  Celikyilmaz, Luke Zettlemoyer, and Veselin Stoyanov.
\newblock Opt-iml: Scaling language model instruction meta learning through the
  lens of generalization.
\newblock {\em ArXiv}, abs/2212.12017, 2022.

\bibitem{Longpre2023TheFC}
S.~Longpre, Le~Hou, Tu~Vu, Albert Webson, Hyung~Won Chung, Yi~Tay, Denny Zhou,
  Quoc~V. Le, Barret Zoph, Jason Wei, and Adam Roberts.
\newblock The flan collection: Designing data and methods for effective
  instruction tuning.
\newblock In {\em International Conference on Machine Learning}, 2023.

\bibitem{CodaForno2023MetaincontextLI}
Julian Coda-Forno, Marcel Binz, Zeynep Akata, Matthew~M. Botvinick, Jane~X.
  Wang, and Eric Schulz.
\newblock Meta-in-context learning in large language models.
\newblock {\em ArXiv}, abs/2305.12907, 2023.

\bibitem{Schuhmann2022LAION5BAO}
Christoph Schuhmann, Romain Beaumont, Richard Vencu, Cade Gordon, Ross
  Wightman, Mehdi Cherti, Theo Coombes, Aarush Katta, Clayton Mullis, Mitchell
  Wortsman, Patrick Schramowski, Srivatsa Kundurthy, Katherine Crowson, Ludwig
  Schmidt, Robert Kaczmarczyk, and Jenia Jitsev.
\newblock Laion-5b: An open large-scale dataset for training next generation
  image-text models.
\newblock {\em ArXiv}, abs/2210.08402, 2022.

\bibitem{Zhu2023MultimodalCA}
Wanrong Zhu, Jack Hessel, Anas Awadalla, Samir~Yitzhak Gadre, Jesse Dodge, Alex
  Fang, Youngjae Yu, Ludwig Schmidt, William~Yang Wang, and Yejin Choi.
\newblock Multimodal c4: An open, billion-scale corpus of images interleaved
  with text.
\newblock {\em ArXiv}, abs/2304.06939, 2023.

\bibitem{tong2022videomae}
Zhan Tong, Yibing Song, Jue Wang, and Limin Wang.
\newblock Video{MAE}: Masked autoencoders are data-efficient learners for
  self-supervised video pre-training.
\newblock In {\em Advances in Neural Information Processing Systems}, 2022.

\bibitem{anonymous2023understanding}
Anonymous.
\newblock Understanding multimodal instruction format for in-context learning.
\newblock In {\em Submitted to The Twelfth International Conference on Learning
  Representations}, 2023.
\newblock under review.

\bibitem{radford2021learning}
Alec Radford, Jong~Wook Kim, Chris Hallacy, Aditya Ramesh, Gabriel Goh,
  Sandhini Agarwal, Girish Sastry, Amanda Askell, Pamela Mishkin, Jack Clark,
  et~al.
\newblock Learning transferable visual models from natural language
  supervision.
\newblock In {\em International conference on machine learning}, pages
  8748--8763. PMLR, 2021.

\bibitem{hu2021lora}
Edward~J Hu, Yelong Shen, Phillip Wallis, Zeyuan Allen-Zhu, Yuanzhi Li, Shean
  Wang, Lu~Wang, and Weizhu Chen.
\newblock Lora: Low-rank adaptation of large language models.
\newblock {\em arXiv preprint arXiv:2106.09685}, 2021.

\bibitem{Zhang2023VideoLLaMAAI}
Hang Zhang, Xin Li, and Lidong Bing.
\newblock Video-llama: An instruction-tuned audio-visual language model for
  video understanding.
\newblock {\em ArXiv}, abs/2306.02858, 2023.

\bibitem{Li2023VideoChatCV}
Kunchang Li, Yinan He, Yi~Wang, Yizhuo Li, Wen Wang, Ping Luo, Yali Wang, Limin
  Wang, and Yu~Qiao.
\newblock Videochat: Chat-centric video understanding.
\newblock {\em ArXiv}, abs/2305.06355, 2023.

\bibitem{Maaz2023VideoChatGPTTD}
Muhammad Maaz, Hanoona Rasheed, Salman Khan, and Fahad~Shahbaz Khan.
\newblock Video-chatgpt: Towards detailed video understanding via large vision
  and language models.
\newblock {\em ArXiv}, abs/2306.05424, 2023.

\bibitem{Luo2023ValleyVA}
Ruipu Luo, Ziwang Zhao, Min Yang, Junwei Dong, Ming-Hui Qiu, Pengcheng Lu, Tao
  Wang, and Zhongyu Wei.
\newblock Valley: Video assistant with large language model enhanced ability.
\newblock {\em ArXiv}, abs/2306.07207, 2023.

\bibitem{rasley2020deepspeed}
Jeff Rasley, Samyam Rajbhandari, Olatunji Ruwase, and Yuxiong He.
\newblock Deepspeed: System optimizations enable training deep learning models
  with over 100 billion parameters.
\newblock In {\em Proceedings of the 26th ACM SIGKDD International Conference
  on Knowledge Discovery \& Data Mining}, pages 3505--3506, 2020.

\bibitem{loshchilov2018decoupled}
Ilya Loshchilov and Frank Hutter.
\newblock Decoupled weight decay regularization.
\newblock In {\em International Conference on Learning Representations}, 2018.

\bibitem{tiny_chat}
Tinychat: Large language model on the edge.
\newblock \url{https://hanlab.mit.edu/blog/tinychat}.
\newblock Accessed: 2023-10-20.

\bibitem{lin2014microsoft}
Tsung-Yi Lin, Michael Maire, Serge Belongie, James Hays, Pietro Perona, Deva
  Ramanan, Piotr Doll{\'a}r, and C~Lawrence Zitnick.
\newblock Microsoft coco: Common objects in context.
\newblock In {\em Computer Vision--ECCV 2014: 13th European Conference, Zurich,
  Switzerland, September 6-12, 2014, Proceedings, Part V 13}, pages 740--755.
  Springer, 2014.

\bibitem{krishna2017visual}
Ranjay Krishna, Yuke Zhu, Oliver Groth, Justin Johnson, Kenji Hata, Joshua
  Kravitz, Stephanie Chen, Yannis Kalantidis, Li-Jia Li, David~A Shamma, et~al.
\newblock Visual genome: Connecting language and vision using crowdsourced
  dense image annotations.
\newblock {\em International journal of computer vision}, 123:32--73, 2017.

\end{thebibliography}

\clearpage
\appendix
\section{Data} \label{sec:data}

\paragraph{Image instruction-following dataset enriched with GCoT response.} \label{data_gcot}

The majority of vision-language tasks can be generally viewed as Visual Question Answering (VQA) tasks, requiring the model to provide answers to queries related to the image. Therefore, we collect 4 VQA datasets to generate GCoT response by ChatGPT, including VQAv2~\cite{goyal2017making}, OK-VQA~\cite{marino2019ok}, GQA~\cite{hudson2019gqa}, and TDIUC~\cite{kafle2017analysis}. Except for GQA, the image source for these tasks is MSCOCO~\cite{lin2014microsoft}, which contains many images, but each image has fewer annotations of caption and object. This may result in the object from the question not having corresponding position information during step (2), making it difficult for ChatGPT to provide an accurate reasoning process. Therefore, we use the Visual Genome dataset~\cite{krishna2017visual} as a supplement, as it has richer annotations and intersects with MSCOCO. The GQA task provides detailed object annotations but lacks captions, which presents a challenge to ChatGPT in comprehending the overall content of the image. So we organize the objects, attributes, and their relationships in the annotations into sentences, which are used to describe the relationships between two objects in the image in place of captions. After preparation, we prompt ChatGPT to follow the aforementioned three steps to generate GCoT templates step by step. The prompts can be found in Table~\ref{tab:gcot_resp}. In addition, we also include LLaVA-instruct-80k~\cite{Liu2023VisualIT} and SVIT~\cite{zhao2023svit} datasets to enhance the model’s instruction-following capability. In summary, in the first stage, \model{} is trained on an image instruction-following dataset comprising approximately 10.7k examples. Within this dataset, there are 9,645 VQA examples accompanied by GCoT responses, which are generated by ChatGPT. \looseness=-1

\paragraph{Video instruction-following dataset based on BDD-X.}

To transition the model's powerful scene understanding and reasoning ability, which has been fine-tuned on the image instruction-following datasets, to the driving video domain, we construct an autonomous driving-related instruction-following dataset based on BDD-X, and at the same time retrieve in-context examples to generate few-shot templates for training to enhance model's in-context learning ability by retrieve method.

\section{Prompts} \label{prompts}

The prompts used to instruct ChatGPT to generate the grounded CoT process with three thinking steps for VQA tasks are shown in Table~\ref{tab:gcot_resp}.

\begin{table*}[!htb]

\begin{tcolorbox}[colback=gray!10]
\begin{tabular}{p{0.97\columnwidth} l}

\textcolor{blue}{\textbf{System Message}} & \\

Give you some captions, each describing the image you are observing and specific object locations within the image are given, along with detailed coordinates. These coordinates are in the form of bounding boxes, represented as (x1, y1, x2, y2) with floating numbers ranging from 0 to 1. These values correspond to the top left x, top left y, bottom right x, and bottom right y. Besides, you will receive a question about the image and the reference answer to this question. & \\
I want you to create a paragraph to explain how to obtain the reference answer for the question. To create the paragraph, follow these steps: & \\
1. Create a short sentence to describe the image by using captions.  & \\
2. Find the question object contained in the question. If the question object exists in the provided objects, create a sentence describing the object and its bounding box; Instead, find the most relevant one with the question object from the provided objects and create a sentence describing the object and its bounding box. & \\
3. If the answer to the question requires reasoning, create a sentence to explain the question. For instance, the question is "When was this piece of sporting equipment invented?" and if in step 3 you have determined that the "sports equipment" is a surfboard, based on your background knowledge, the sentence you need to create is: "In 1926 an American surfer named Tom Blake (1902 - 1994) invented the very first, hollow surfboard."; If the answer to the question does not require reasoning and can be derived directly by looking at the image, for example, the question "What color of the bear is?", then in this step, you do not need to create any sentences to explain it. & \\
Note: You need to determine the position between the objects based on their bounding box. For example object A: [x1, y1, x2, y2], object B: [a1, b1, a2, b2]. If x2 $>$ a2, it means that object A is to the right of object B, and vice versa. If y2 $>$ b2, it means that object A is on top of object B, and vice versa. & \\

\hrulefill & \\

\textcolor{blue}{\textbf{User}} & \\
Captions: & \\
Man in all black swimsuit walking down a beach with his surfboard. & \\
A man in a wetsuit carrying a surfboard to the water. & \\
A person with a surfboard walking on a beach. & \\
A person with a surfboard walks to the water.  & \\
A man carrying a surfboard across a sandy beach. & \\
& \\
Objects: & \\
bird: [0.095, 0.797, 0.355, 0.849],
surfboard: [0.388, 0.418, 1.254, 0.977],
person: [0.431, 0.222, 0.941, 1.362] & \\
& \\
Question: When was this piece of sporting equipment invented? & \\
Reference Answer: 1926 & \\

\hrulefill & \\
\textcolor{blue}{\textbf{Output}} & \\

1. The picture shows a man carrying a surfboard across a sandy beach. & \\
2. So the sporting equipment in question should refer to the surfboard. & \\
3. In 1926 an American surfer named Tom Blake (1902 - 1994) invented the very first, hollow surfboard. & \\

\end{tabular}
\end{tcolorbox}

\caption{One example to prompt ChatGPT for generating GCoT to enhance the VLM's capabilities of fine-grained multimodal understanding and reasoning, which are considered the important proficiency in ADs.\looseness=-1}

\label{tab:gcot_resp}

\end{table*}

\end{document}